\documentclass[]{interact}

\usepackage{epstopdf}

\usepackage{natbib}
\bibpunct[, ]{(}{)}{;}{a}{}{,}
\renewcommand\bibfont{\fontsize{10}{12}\selectfont}

\usepackage{amsmath}    
\usepackage{subcaption} 
\usepackage{booktabs} 
\usepackage{lmodern}    
\usepackage{setspace}   

\theoremstyle{plain}

\theoremstyle{definition}

\theoremstyle{remark}

\begin{document}

\articletype{This is an Accepted Manuscript of an article published by Taylor \& Francis in International Journal of Remote Sensing on 03 Dec 2025. The published final Version of Record is openly available at: https://doi.org/10.1080/01431161.2025.2579803.
}
\title{Towards autonomous photogrammetric forest inventory using a lightweight under-canopy robotic drone}


\author{
\name{V\"{a}in\"{o} Karjalainen\textsuperscript{a,$^\ast$}\thanks{$^\ast$ Authors contributed equally. CONTACT: V\"{a}in\"{o} Karjalainen. Email: vaino.karjalainen@nls.fi, Address: Department of Remote Sensing and Photogrammetry,
Finnish Geospatial Research Institute FGI, The National Land Survey of Finland, Vuorimiehentie 5, FI-02150, Espoo, Finland}, Niko Koivum\"{a}ki\textsuperscript{a,$^\ast$}, Teemu Hakala\textsuperscript{a}, Jesse Muhojoki\textsuperscript{a}, Eric Hyypp\"{a}\textsuperscript{a}, Anand George\textsuperscript{a}, Juha Suomalainen\textsuperscript{a}, Eija Honkavaara\textsuperscript{a}}
\affil{\textsuperscript{a}Department of Remote Sensing and Photogrammetry, Finnish Geospatial Research Institute FGI, The National Land Survey of Finland, Espoo, Finland}
}

\maketitle

\begin{abstract}
Drones are increasingly used in forestry to capture high-resolution remote sensing data, supporting enhanced monitoring, assessment, and decision-making processes. While operations above the forest canopy are already highly automated, flying inside forests remains challenging, primarily relying on manual piloting. In dense forests, relying on the Global Navigation Satellite System (GNSS) for localization is not feasible. In addition, the drone must autonomously adjust its flight path to avoid collisions. Recently, advancements in robotics have enabled autonomous drone flights in GNSS-denied obstacle-rich areas. In this article, a step towards autonomous forest data collection is taken by building a prototype of a robotic under-canopy drone utilizing state-of-the-art open source methods and validating its performance for data collection inside forests. Specifically, the study focused on camera-based autonomous flight under the forest canopy and photogrammetric post-processing of the data collected with the low-cost onboard stereo camera. The autonomous flight capability of the prototype was evaluated through multiple test flights in boreal forests. The tree parameter estimation capability was studied by performing diameter at breast height (DBH) estimation. The prototype successfully carried out flights in selected challenging forest environments, and the experiments showed promising performance in forest 3D modeling with a miniaturized stereoscopic photogrammetric system. The DBH estimation achieved a root mean square error (RMSE) of 3.33 - 3.97 cm (10.69 - 12.98 \%) across all trees. For trees with a DBH less than 30 cm, the RMSE was 1.16 - 2.56 cm (5.74 - 12.47 \%). The results provide valuable insights into autonomous under-canopy forest mapping and highlight the critical next steps for advancing lightweight robotic drone systems for mapping complex forest environments.
\end{abstract}


\section{Introduction}

In recent decades, remote sensing measurements within forests have been increasingly used to capture detailed forest information. Terrestrial laser scanning (TLS) made a breakthrough in the early 2000s, followed by the development of various solutions, including static laser scanning using tripods \citep{bienert2007tree} and mobile solutions, e.g., handheld devices \citep{ryding2015assessing, bauwens2016forest, marselis2016deriving, balenovic2021hand}, backpacks \citep{liang2014possibilities, liang2018situ, hyyppa2020accurate}, ground-based vehicles \citep{forsman2016tree, liang2018situ, bienert2018comparison} or 
drones \citep{chisholm2013uav, brede2017comparing, wieser2017case, hyyppa2020under}. Similarly, the use of structure-from-motion (SfM) based point clouds for forest mapping has been demonstrated using different camera solutions \citep[e.g.,][]{liang2014use,wallace2016assessment,kuvzelka2018mapping,piermattei2019terrestrial}. 
Among these methods, drone-based solutions offer several benefits, including the ability to move in areas with challenging terrain and the high potential for automatization. 

Hence, the use of drones in forest research has increased rapidly and offers a practical solution to collect forest data above the forest canopy \citep{puliti2015forestUAS, tuominen2017hyperspectral}. Modern commercial drones are highly automated and can execute data collection missions highly autonomously in open-air environments using autopilot systems. However, their positioning systems typically rely exclusively on GNSS signals, limiting autonomous flights to open areas where these signals are reliable. In dense forests, GNSS-based localization becomes unreliable due to signal blockages and multipath effects caused by trees \citep{gnss_multipath}. Furthermore, navigating within forests requires the ability to autonomously avoid obstacles such as trees and bushes along the flight path. Consequently, drone data collection flights under the forest canopy are still primarily controlled manually by human pilots \citep[e.g.,][]{hyyppa2020under, liang2019forest, krisanski2020photodrone, prabhu2024treeFalcon}.

Recent advances in robotics and the enhanced computational capabilities of small embedded computers have enabled drone autonomy even in cluttered GNSS-denied environments. Although technology is still in the early stage of development, some recent studies have proposed various methods that have been successfully tested in forest environments \citep[e.g.,][]{LiuEtAl, loquercio2021learning, zhou2022swarm}. However, the proposed methods are typically validated by reporting a single successful test flight in a specific forest environment. Most studies assess the effects of varying object density only in simulators, with only a couple of methods tested through multiple test flights inside real forests. 

Moreover, most of the studies proposing methods for autonomous under-canopy drone trajectory planning have the scope solely on aerial robotics, and the sensor data is primarily utilized only for navigation. Only a few pioneering studies have yet explored the potential of custom robotic under-canopy drone systems in forestry applications \citep{LiuEtAl, liang2024forest, cheng2024Treescope}. Notably, all of these studies have utilized Ouster OS1 3D LiDAR (Ouster, San Francisco, CA, USA) as the main sensor. Whereas LiDARs have their strengths, such as 360\textdegree{} horizontal field of view (FOV), long sensing range, and robustness against varying illumination, they tend to be heavier and have higher power consumption than cameras \citep{sensors}. In the case of drone development, where optimizing the weight and battery consumption plays a key role, these factors lead to larger and heavier drone platforms and flight batteries that require more space between obstacles to fly safely inside the forest. Furthermore, LiDARs with suitable quality are more expensive than camera systems with built-in 3D imaging capability, leading to increased building cost. The data collected with the LiDARs is also limited to point clouds only, whereas the images collected with cameras can also be used for image-based analysis.

Hence, although some LiDAR-based robotic drone systems have been successfully applied to under-canopy mapping of forest parameters, the potential of smaller and cheaper camera-based systems remains largely unexplored. Furthermore, the robotic drone systems proposed in the literature are typically not evaluated through multiple flight tests in different forest environments, leaving the reliability and suitability to diverse forest environments as an open question. 
 
Therefore, in this study, a prototype of a small camera-based robotic drone system that can fly autonomously in challenging GNSS-denied forests was built and developed using custom hardware and state-of-the-art robotic methods. The study aimed to tackle the research gaps in flight reliability of camera-based robotic drone systems in various Boreal forest environments, as well as the potential of photogrammetric DBH estimation with autonomous drones in such environments. The performance and the reliability of autonomous flight capabilities were evaluated by performing multiple test flights in Boreal forests with varying difficulty. In addition, the image data collected with the low-cost onboard stereo camera were processed into point clouds, and the DBH values were estimated for the trees detected from the dataset. To the best of our knowledge, this study is the first to empirically validate a miniaturized, autonomous, camera-based robotic drone for under-canopy forest reconstruction.

Specifically, the study focused on three research questions. What level of autonomous flight reliability can be achieved in challenging boreal forests with current methods for camera-based robotic drone navigation? What level of DBH estimation accuracy can be achieved with the low-cost onboard stereo camera data recorded during the flight? What are the remaining bottlenecks to be tackled in follow-up studies in order to achieve large-scale autonomous photogrammetric forest data collecting with miniaturized under-canopy robotic drone systems?

The main contributions of this study can be highlighted as follows:
\begin{itemize}
    \item Build and development of a lightweight robotic drone prototype capable of flying within GNSS-denied forest environments, using custom hardware and state-of-the-art robotic algorithms. The reliability of the system was comprehensively evaluated through multiple test flights in boreal forests with varying tree densities and difficulty levels. 
    \item Combining autonomous flight capabilities to under-canopy photogrammetric forest reconstruction by processing the data from a low-cost onboard stereo camera to 3D point clouds. 
    \item Validation of autonomous flight data collection based on stereo cameras for the estimation of forest parameters through DBH estimation, including an evaluation of the impact of various factors on the accuracy of the estimate.
    \item Comprehensive discussion of current limitations to large-scale, fully autonomous photogrammetric under-canopy forest monitoring, along with proposed solutions and further research directions.
\end{itemize}

The rest of this study is structured as follows. Section \ref{sec:relatedWork} reviews the latest studies that estimate DBH using drones flying inside forests and recent open source solutions for autonomous drone flights inside forests. Section \ref{sec:methods} describes the algorithms, hardware, experiments, and methods utilized in this study. Experimental results are presented in Section \ref{sec:Results}. Section \ref{sec:Discussion} discusses the results and potential improvements and concludes the study.

\section{Related work}
\label{sec:relatedWork}

\subsection{DBH estimation from under-canopy drone data}
\label{sec:relatedEstimation}

In a forest inventory, DBH is one of the most important attributes to estimate, as it provides a measure of the size of a tree and enables the prediction of the total volume of the stem alone or in combination with the three heights \citep{laasasenaho1982taper}.

In the existing literature, LiDAR-based methods have been extensively studied to efficiently and accurately estimate various tree attributes, such as DBH, tree height, and stem volume. To date, several LiDAR-based mobile platforms have been shown to enable DBH estimation of individual trees with an RMSE of 1-2 cm \citep{ryding2015assessing, bauwens2016forest,cabo2018comparing, chen2019applicability, hyyppa2020accurate, hyyppa2020under, hyyppa2022direct}, though most studies have focused on relatively sparse forests with little understory vegetation. 

\citet{chisholm2013uav} were the first to demonstrate DBH measurements from point cloud data collected with a manually piloted drone flying under the forest canopy. The utilized custom-built drone system included a Hokuyo UTM-30LX 2D LiDAR (Hokuyo, Osaka, Japan). 

\citet{hyyppa2020under} collected high-quality point cloud data using a Kaarta Stencil-1 (Kaarta, Pittsburgh, Pennsylvania, USA) laser scanner mounted horizontally at the bottom of a manually piloted drone in two plots of varying complexity in a boreal forest with a tree density of 410-420 trees/ha. Importantly, their stem diameter estimation algorithm utilized the temporal order of data points to detect stem arcs, thus mitigating distortions caused by slow temporal drifts of the drone trajectory. In a follow-up study, \citet{hyyppa2020comparison} compared various mobile laser scanning methods, including backpack, handheld, Under-Canopy drone, and Above-Canopy drone, for forest data collection. In this study, the same manually piloted under-canopy drone system was utilized in the same forest plots as in \citet{hyyppa2020under}.

\cite{wang2021seamless} combined data from above and under canopy flights of a manually piloted drone equipped with a Riegl miniVUX laser scanner (RIEGL Laser Measurement Systems GmbH, Horn, Austria). The study area was a boreal forest plot with an approximate tree density of 200 trees/ha.

\citet{muhojoki24BENCHMARKING} compared the accuracy of the estimated tree attributes of several LiDAR-based mobile systems, including two commercial under-canopy drone systems, a Hovermap system (2020 version, Emesent, Milton, Australia) equipped with a Velodyne VLP-16 laser scanner (Velodyne Lidar, San Jose, CA, USA) and a Deep Forestry system (Deep Forestry, Uppsala, Sweden) utilizing an Ouster OS0-32 Rev. 5 laser scanner (Ouster, San Francisco, CA, USA). 
The Deep Forestry drone was flown manually, but the Hovermap drone system was able to fly autonomously in sparse forests where the distance between trees was at least 3 m. Both drone systems were used to collect point clouds under the canopy of three sparse forest plots. 

In addition to manually controlled drones and large autonomous commercial drone systems, recently a few pioneering studies have applied smaller custom robotic drone platforms with a high automation level for LiDAR-based under-canopy DBH estimation. \citet{liang2024forest} applied an autonomously operated under-canopy drone system to collect laser scanning data inside a 50 m $\times$ 30 m forest plot with a tree density of approximately 1000 trees/ha. To enable autonomous flight, a predefined global path was planned, in which the drone system made real-time adjustments during the flight according to real-time environmental conditions.  

\citet{LiuEtAl} employed a LiDAR-based semantic lidar odometry and mapping framework (SLOAM) \citep{chen2020sloam} algorithm to calculate the number of trees in the explored area. Although in this study autonomous flight was demonstrated without the SLOAM algorithm and tree number calculation was only demonstrated with one manually piloted test flight, the same drone system has later been used in various follow-up studies for tree parameter estimation. \citet{prabhu2024treeFalcon} extended the system to have 3D trajectory planning capability and used the platform to demonstrate metric-semantic map merging between multiple robotic drones, as well as DBH and tree trunk diameter profile estimation with the onboard LiDAR data.
However, DBH data collection flights were performed in manual flight mode. \citet{cheng2024Treescope} used the same drone platform to collect a dataset for tree mapping based on LiDAR. In addition to data from orchards, the dataset contains the onboard LiDAR data from two forest areas. Although the drone was manually piloted in one of these forests, data collection in the other forest consisting of loblolly pine trees was carried out in autonomous flight mode. They also estimated DBH values for the LiDAR data collected during both autonomous and manual flights. 



During the past decade, photogrammetric systems producing SfM point clouds have been suggested to provide an efficient, low-cost alternative to traditional field measurements of tree attributes, such as DBH. Although several studies have investigated terrestrial photogrammetric systems (e.g., \citet{liang2014use,surovy2016accuracy, mokrovs2018high,piermattei2019terrestrial, xu2023sfm_vs_ls, gao2022automaticDBH}) and above-canopy photogrammetric systems (e.g., \citet{wallace2016assessment, ye2019analysing}) for tree attribute estimation, only relatively few studies have tested the use of under-canopy drone systems for the collection of SfM point clouds enabling tree attribute measurements \citep{kuvzelka2018mapping, krisanski2020photodrone,  shimabuku2023diameter, he2025estimating360}.

\citet{kuvzelka2018mapping} evaluated two manually operated drone systems, a DJI Phantom 4 Pro (DJI, Shenzhen, China) and a DJI Mavic Pro (DJI, Shenzhen, China), and various analysis methods to estimate DBH from photogrammetric point clouds. The systems were evaluated in two sparse forest plots with tree densities of 270-290 trees/ha. 

\citet{krisanski2020photodrone} performed manual flights with a DJI Phantom 4 Pro in two forest plots with stem densities of 885 trees/ha and 980 to acquire high-density SfM point cloud data and estimated the DBH of individual trees from the point clouds.

\citet{shimabuku2023diameter} performed DBH estimation on two small subtropical forest plots using SfM point clouds acquired with an under-canopy flying DJI Mini 3 Pro drone (DJI, Shenzhen, China). The drone system was able to detect obstacles in the horizontal direction, but needed manual piloting to avoid obstacles in the vertical direction and to guide the drone back to the intended flight path after automatic obstacle avoidance in the horizontal direction.

Recently, \citet{he2025estimating360} mounted a 360\textdegree{} FOV Insta360 camera (Arashi Vision, Shenzhen, China) as an additional payload to the same LiDAR-based autonomous drone platform used in \citet{LiuEtAl} and post-processed the photogrammetric data to 3D point clouds. However, in the study, it was not mentioned whether the flights were performed in autonomous flight mode or with manual pilot mode.

The RMSEs of the DBH estimates in the relevant studies are presented in Table \ref{tab:reference_DBHestimation_stats}. From the studies where multiple analysis methods were evaluated, the lowest RMSEs are presented. The study by \citet{he2025estimating360} was omitted from the table, as the study reported only the median relative error (11.5 \%) of the accuracy of the DBH estimation instead of the RMSE. Although some of the relevant studies also estimated tree volumes by flying the flight path several times with different altitudes, with rotating sensors, or with LiDARs with high vertical FOV, only DBH estimation accuracies are included in the table.

\begin{table*}[!h]
\caption{Accuracy of the stem diameter breast height (DBH) estimations in other studies.  SfM-MVS: Structure-from-motion and multi-view stereo; ULS: Under canopy laser scanning; ALS: Above canopy laser scanning. *averaged values from three plots. The study by \citet{he2025estimating360} was omitted from this table since they reported only median errors instead of RMSE. }
\begin{center}
    \resizebox{\textwidth}{!}{%
  \begin{tabular}{lcccccc}
  \hline 
    \textbf{Dataset} & \textbf{Type}& \textbf{Number of trees} & \textbf{RMSE (cm)} & \textbf{RMSE \%} & \textbf{Bias (cm)} & \textbf{Bias \%}\\
    \hline 
    \vtop{\hbox{\strut{Chisholm et al.}}\hbox{\strut{(2013)}}} & ULS & 12 & - & 25.1 & - & -\\
    \vtop{\hbox{\strut{Liang et al.}}\hbox{\strut{(2024) traj. I}}} & ULS & 143 & 9.79 & 61.41 & 7.18 & 45.03\\
    \vtop{\hbox{\strut{Liang et al.}}\hbox{\strut{(2024) traj. II}}} & ULS & 143 & 5.13 & 22.01 & 3.40 & 14.62\\
    \vtop{\hbox{\strut{Hyypp\"{a} et al.}}\hbox{\strut{(2020a) plot I}}} & ULS & 42 & 0.69 & 2.2 & 0.31 & 1.1\\
    \vtop{\hbox{\strut{Hyypp\"{a} et al.}}\hbox{\strut{(2020a) plot II}}} & ULS & 43 & 0.92 & 3.1 & 0.29 & 1.0\\
    \vtop{\hbox{\strut{Hyypp\"{a} et al.}}\hbox{\strut{(2020c) plot I}}} & ULS & 42 & 0.6 & 2.3 & 0.34 & 1.3\\
    \vtop{\hbox{\strut{Hyypp\"{a} et al.}}\hbox{\strut{(2020c) plot II}}} & ULS & 43 & 1.1 & 3.5 & 0.12 & 0.4\\
    \vtop{\hbox{\strut{Wang et al.}}\hbox{\strut{(2021)}}}  & ULS\&ALS & - & 7.95 & 33.35 & 3.72 & 15.61\\
    \vtop{\hbox{\strut{Muhojoki et al.}}\hbox{\strut{(2024)}}}*  & ULS & 50 & 1.50-1.60 & 6.70-6.90 & 0.10-0.70 & 0.60-3.00\\
    \vtop{\hbox{\strut{Prabhu et al.}}\hbox{\strut{(2024)}}}  & ULS & 20 & 1.67 & 7.42 & - & -\\
    \vtop{\hbox{\strut{Cheng et al.}}\hbox{\strut{(2024) Plot I}}}  & ULS & 77 & 3.8 & 12.8 & - & -\\
    \vtop{\hbox{\strut{Cheng et al.}}\hbox{\strut{(2024) Plot II}}}  & ULS & 20 & 2.0 & 7.5 & - & -\\
    \vtop{\hbox{\strut{Shimabuku et al.}}\hbox{\strut{(2023) plot I}}} & Drone SfM-MVS & 18 & 0.7 & 3.6 & - & - \\
    \vtop{\hbox{\strut{Shimabuku et al.}}\hbox{\strut{(2023) plot II}}} & Drone SfM-MVS & 7 & 0.4 & 3.2 & - & - \\
    \vtop{\hbox{\strut{Ku{\v{z}}elka and}}\hbox{\strut{Surov{\`y} (2018)}}} & Drone SfM-MVS & 119 & 2.63-5.31 & 7.01-13.3 & 0.27-3.77 & 0.69-10.11\\
    \vtop{\hbox{\strut{Krisanski et al.}}\hbox{\strut{(2020)}}} & Drone SfM-MVS & 29 & 4.1 & - & - & -\\
    \hline
\end{tabular}  
}
\label{tab:reference_DBHestimation_stats}
\end{center}
\end{table*}

In summary of the related work, most of the studies for under-canopy drone-based DBH estimation have been conducted with manually piloted drones. Although a few studies have explored the potential of custom robotic drone systems capable of flying without commands from the pilot during the flight, all of these systems have been based on LiDARs. All of the studies based on photogrammetry and SfM point clouds have utilized commercial drone systems, which have required manual piloting during data collection. 

The highest estimation accuracies have been obtained with large systems unable to fly in dense forests or with test plots containing very few reference trees, but miniaturized systems have still yielded promising results. Furthermore, all studies utilizing SfM point clouds have been performed in environments that differ significantly from boreal forests.


\subsection{Autonomous flying inside forests }
\label{sec:relatedAutonomous}

Even though only a few studies have yet utilized robotic under-canopy drones in DBH estimation, drone autonomy itself has been an actively researched topic in the robotics community during the past few years. Recent advances in mobile robotics and the improved computational capabilities of small embedded computers have made it realistic to build drones capable of real-time trajectory planning in cluttered GNSS-denied environments. Although technology is still in the early stages of development, some recent studies have proposed various methods that have been successfully tested in forest environments. Most importantly, the authors of these methods have also shared their source codes to make them openly available to the research community to encourage more development in the field. Although the methods utilize various sensor setups, they can be divided into LiDAR-based methods \citep{LiuEtAl, prabhu2024treeFalcon, ren2022BubblePlanner, LiuIPC2024, Ahmad2022Bioinspired} and camera-based methods \citep{CamposEtAl,zhou2022swarm, loquercio2021learning, nguyen2023Unceratainity} based on the sensor utilized for the detection of obstacles in the flight path. Since GNSS is not a suitable method for localization under the forest canopy, these methods rely on various Simultaneous localization and mapping (SLAM) or odometry algorithms and onboard sensor data to obtain the pose estimate in a local coordinate system. Most of these methods tackle the problem of collision-free trajectory planning using 3D grid maps, but the solutions by \citet{loquercio2021learning} and \citet{nguyen2023Unceratainity} replaced individual mapping and trajectory planning subtasks by deep neural networks mapping the depth images directly to collision-free trajectories. In addition, the methods proposed by \citet{zhou2022swarm} and \citet{Ahmad2022Bioinspired} concentrated a swarm of drones instead of the autonomous flying of an individual drone. 

However, most of these methods were validated by a single successful autonomous test flight in a specific forest environment. Only four of the articles reported multiple test flights inside real forests \citep{loquercio2021learning, ren2022BubblePlanner, LiuIPC2024, prabhu2024treeFalcon}. From these, \citep{loquercio2021learning} and \citep{prabhu2024treeFalcon} were the only methods tested at least in two different forest types. Furthermore, since the methods by \citep{loquercio2021learning}, \citep{ren2022BubblePlanner}, and \citep{LiuIPC2024} were all targeting high-speed flight, the test forests reported in these studies were relatively sparse, with a low amount of low-hanging branches and understory vegetation. The autonomous trajectory planning method by \citet{prabhu2024treeFalcon} was also tested only in environments where the space between the trees was larger than in typical forests, and the data-collecting flights in denser forests were manually piloted. 

Therefore, open questions remain regarding the applicability and reliability of these algorithms in diverse forest environments, especially challenging boreal forests.

\section{Materials and methods}
\label{sec:methods}

\subsection{Autonomous navigation algorithms} \label{sec:droneNav}

In this study, the objective was to develop a prototype lightweight camera-based drone capable of avoiding obstacles autonomously under the forest canopy and operating without GNSS. The drone was required to fly autonomously from the takeoff location to a specified goal point while recording stereo camera data during the flight. This prototype will later serve as a development platform for more extended and sophisticated data collection tasks. 

Among the open source solutions presented in Section \ref{sec:relatedAutonomous}, the solution by \citet{zhou2022swarm} was chosen as the basis of the prototype. The solution was the only open source camera-based solution that was proven to work with only one stereo camera, making it possible to build a smaller and lighter platform. Furthermore, the test environment reported in the original article (dense bamboo forest) was the most difficult among the proposed solutions, although it had a highly different structure from typical boreal forests.  

\begin{figure*}[ht!]
\begin{center}
		\includegraphics[width=1.0\textwidth]{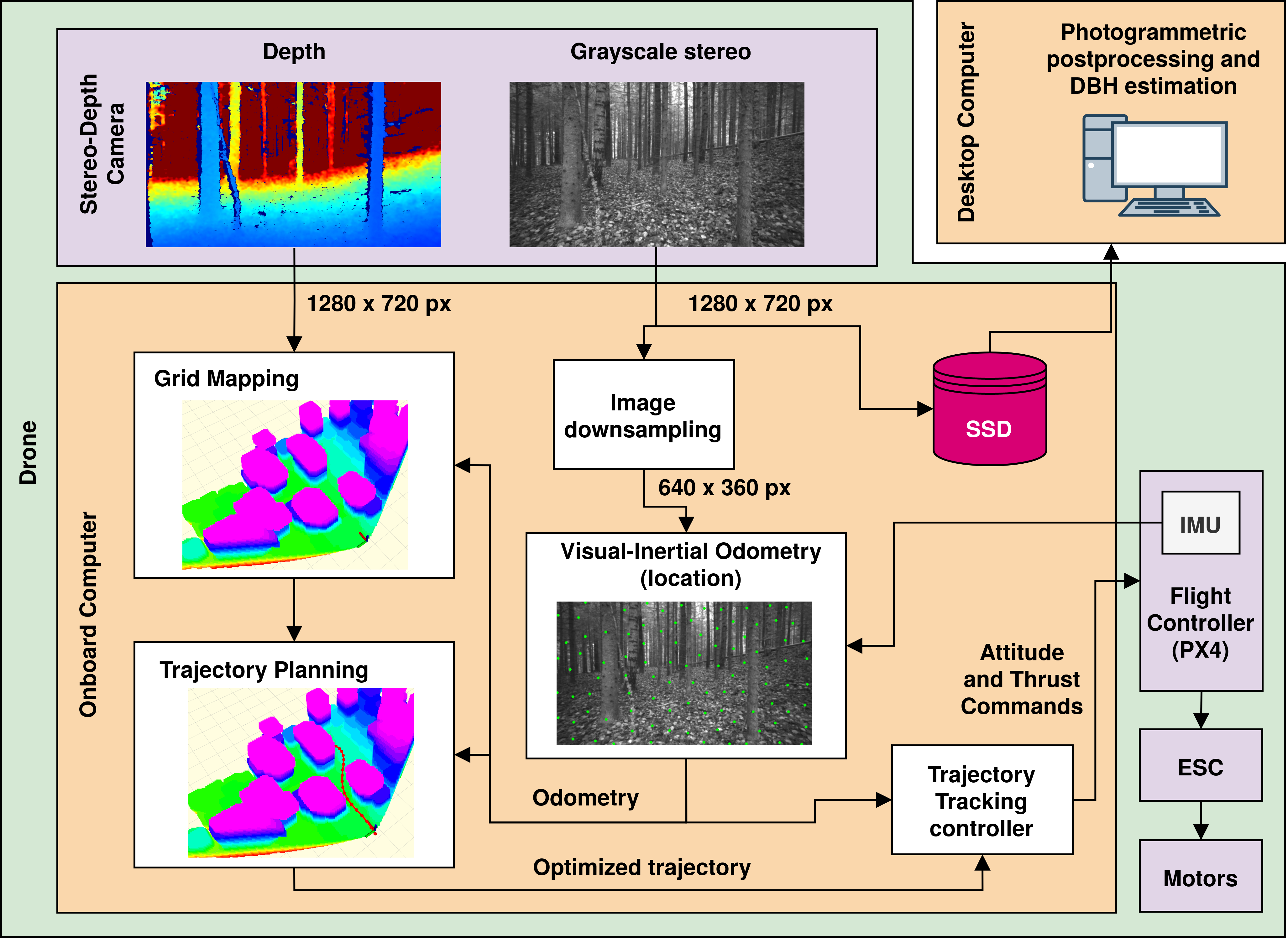}
        \caption{ The system architecture used in the data collection. In comparison to the original EGO-Planner-v2 article by \citet{zhou2022swarm}, UWB-based Drift correction and Wireless trajectory broadcast network modules used for swarm navigation were omitted in this study. Stereo images were downsampled for VIO to guarantee real-time estimation. Full-resolution stereo images were also saved and post-processed to point clouds used in the DBH estimation. }
\label{fig:architecture}
\end{center}
\end{figure*}

The main details of the algorithms used in the chosen solution are briefly described in the following, and the complete system architecture of the prototype is presented in Figure \ref{fig:architecture}. For more detailed information, the readers are referred to the cited original sources.

\subsubsection{EGO-Planner-v2}

The source code of the planner algorithm by \citet{zhou2022swarm} was named on GitHub as EGO-Planner-v2 \citep{EGO-Planner-code}. Although EGO-Planner-v2 is designed for swarm flying, in this study, the algorithm was modified for a single drone system to focus on navigation performance in challenging forest environments. The planner runs on top of the Robot Operating System (ROS) \citep{quigley2009ros}. The mapping of EGO-Planner-v2 is based on probabilistic occupancy grid maps \citep{moravec1985high}. The mapping module uses depth images as input and the system adopts the fixed-size circular buffers proposed by \citet{usenko2017} to maintain the local map. The maps also have a virtual floor and a virtual ceiling defining the minimum and maximum altitudes of the grid map, respectively. The virtual floor and the virtual ceiling restrict the altitude at which the algorithm can plan the paths. 

In trajectory planning, EGO-Planner-v2 uses a MINCO (minimum control) \citep{MINCO} trajectory representation designed for differentially flat systems, such as quadrotors \citep{differentialFlatness}. The planned trajectories are represented as piece-wise polynomial splines with decoupled temporal and spatial parameters \citep{zhou2022swarm}. The trajectories are optimized iteratively by minimizing the weighted sum of metrics defined for trajectory smoothness and flying time. The feasibility of the planned trajectory is forced by deforming the shape to avoid obstacles and by restricting the magnitudes of the trajectory velocity, acceleration, and jerk to meet the dynamical constraints of the drone. For computational efficiency, trajectory constraints are enforced through integrals of penalty functions with large penalty weights, and the integrals are evaluated by a finite sum of equally spaced samples along the timeline. In a simplified form, the optimization of time-dependent objectives $J$ in one piece of the trajectory can be represented by   
\begin{equation}
       {\min} \sum_{x}\lambda_x J_x 
\end{equation}
where $J_x$ are the penalty terms and $\lambda_x$ are the corresponding weights. Subscripts $x=\{s,t,d,o,u\}$ stand for the trajectory smoothness, total flying time, dynamical feasibility, obstacle avoidance, and the uniform distribution of constraint evaluation points along the trajectory, respectively. In addition to the above-mentioned penalty terms,
EGO-Planner-v2 also includes penalty terms for the swarm operations, but those are omitted here. A* is utilized as the global path-search algorithm given as input for the optimization algorithm.

The trajectory planning is continuously performed within a defined local planning distance towards a user-defined global goal until the goal is reached. Since the safety constraints are modeled as integrals in the penalty functions, the feasibility of the trajectory needs to be verified with a post-check after planning \citep{zhou2022swarm}. Due to the non-static probabilistic map, the post-check after trajectory planning guarantees feasibility only at a certain moment. Hence, a collision-check process is continuously running in the background. If that process detects a potential collision, a trajectory replanning is activated immediately. If the replanning fails and the predicted time to collision is under a given threshold, the planner performs an emergency stop. After the drone has stopped, the trajectory planner tries to start up again and find a new feasible trajectory.

\subsubsection{VINS-Fusion}

The Visual-inertial odometry (VIO) algorithm used by \citet{zhou2022swarm} was VINS-Fusion \citep{qin2018vins, qin2019general}(VINS stands for 'Visual-Inertial system'), which used grayscale stereo images and IMU data to track the position and orientation of the drone. 

VINS-Fusion detects Shi-Tomasi corner features \citep{shiTomasi1994}, and the features are tracked between frames with Kanade–Lucas–Tomasi (KLT) Tracker \citep{lucas1981iterative}. IMU data is used for preintegration, i.e., position, velocity, and orientation are integrated from IMU measurements between two frames. The state optimization is performed within a sliding window, including the position, velocity, rotation, and IMU biases. To reduce the computational complexity, old measurements are marginalized into a prior of the estimation process. Certain frames are also selected as keyframes, which are saved to a global pose graph after being marginalized out from the sliding window. Keyframes are selected by considering the number of tracked features and the average parallax between these features in the most recent camera frame and the latest keyframe. 

VINS-Fusion supports loop detection-based relocalization and global pose graph optimization to reduce the accumulated drift in the pose estimate. The loops are detected by storing additional corner features as BRIEF descriptors \citep{BRIEF} to a DBoW2 \citep{DBoW2} bag-of-word database. After a loop detection, the relocalization process aligns the latest window to past poses and optimizes it utilizing also the feature correspondences with past poses. In addition to sliding-window relocalization, the global pose graph is also optimized after loop detections to enforce the global consistency of the poses. Since the roll and pitch angle estimates are drift accumulation-free due to full observability from the gravity measurement of the IMU, the pose graph optimization is performed in 4 degrees of freedom.   

VINS-Fusion also adopts an online temporal calibration method for inertial and visual measurements \citep{onlineTd}. The method can be used to adjust the time offset $t_d$ between the camera and the IMU online. 

In the swarm operations of the original article \citep{zhou2022swarm}, the drones used also relative distances measured with Ultra Wide Band (UWB) sensors to correct the drift in the pose estimate of VINS-Fusion. However, with a single drone system in this study, that function is omitted.

The parameters used in the configuration of VINS-Fusion were chosen so that real-time performance was guaranteed. In the initial tests, the IMU noise values used during the calibration were over-optimistic for VINS-Fusion, causing instability to the estimates, so the IMU noise values were increased further. All VINS-Fusion configuration parameters are listed in Table \ref{tab:VINS-parameters}.

\begin{table}[!h]
\caption{VINS-Fusion parameters.}
\begin{center}
  \begin{tabular}{lc}
  \hline 
   \textbf{Parameter} & \textbf{Value}\\
    \hline 
    Image resolution & $640 \times 360$\\
    Max number of tracked features & 150 \\
    Min distance between features & 40 px \\
    RANSAC threshold & 1 px \\
    Max solver iteration time & 0.04 ms \\
    Max solver iterations & 8 \\
    Keyframe parallax threshold & 10 px \\
    Accelerometer noise SD & 0.2 \\
    Gyroscope noise SD & 0.05 \\
    Accelerometer random walk SD & 0.002 \\
    Gyroscope random walk SD & 0.0005 \\
    \hline
\end{tabular}  
\label{tab:VINS-parameters}
\end{center}
\end{table}

\subsubsection{Tracking controller}

The original EGO-Planner-v2 article \citep{zhou2022swarm} did not specify the method for trajectory tracking. In this study, an open source tracking controller \citep{px4ctrl} by Zhejiang University was used. The tracking controller gets setpoints for position, velocity, acceleration, and yaw from the trajectory planner and converts those to attitude and thrust setpoints. The setpoints are forwarded to the PX4 flight controller software \citep{PX4code} responsible for the low-level attitude and angular rate controls.

\subsection{Drone hardware}\label{sec:Hardware}

In the physical robot drone prototype, an onboard computer NVIDIA Jetson Orin NX (Santa Clara, California, USA) was used. The used PX4-compliant autopilot was Holybro Kakute H7 (Hong Kong, China), which also contained the IMU. The navigation algorithms mentioned in the previous subsection ran on top of the ROS, and the MAVROS protocol was used for communication between the onboard computer and the autopilot. As a camera, Intel RealSense D435 \citep{realsense}\citep{realsense_article} was used. The camera was capturing both depth images for 3D mapping and grayscale stereo images for VIO.

The RealSense D435 onboard camera was configured to the maximum image resolution ($1280 \times 720$) and the update frequency (30 Hz). The exposure value was set to use the built-in automatic exposure control. The stereo images for VIO were downsampled at a resolution of $640 \times 360$ to ensure real-time performance. 

Stereoscopic RealSense D435 has an infrared laser projector emitting dots, improving the triangulation in low light conditions and for homogeneous surfaces. However, the dots of the laser emitter cause problems for VIO since they are moving with the camera. Therefore, in this study, the infrared emitter was disabled. Since sunlight contains infrared light, which interferes with the artificial features emitted \citep{IRoutdoor}, and forests typically do not contain homogeneous surfaces, disabling the emitter should not affect the camera performance in outdoor conditions.

The camera-IMU system was calibrated with an open source camera calibration toolbox, kalibr \citep{kalibr3,kalibr2,kalibr1}. In the calibration, the intrinsic and extrinsic parameters of the left and right grayscale cameras and the time shift between the camera and the autopilot IMU were estimated from a recorded data sequence, where Aprilgrid \citep{apriltag} was used as a calibration target. The IMU noise parameters needed for the calibration were first obtained with the open source tool Allan Variance ROS \citep{AllanVarianceROS}, and the values obtained were multiplied by 10 before the calibration as suggested in the wiki of the kalibr project \citep{kalibrWikiIMU}.

\begin{figure}[h!]
\begin{center}
		\includegraphics[width=1.0\columnwidth]{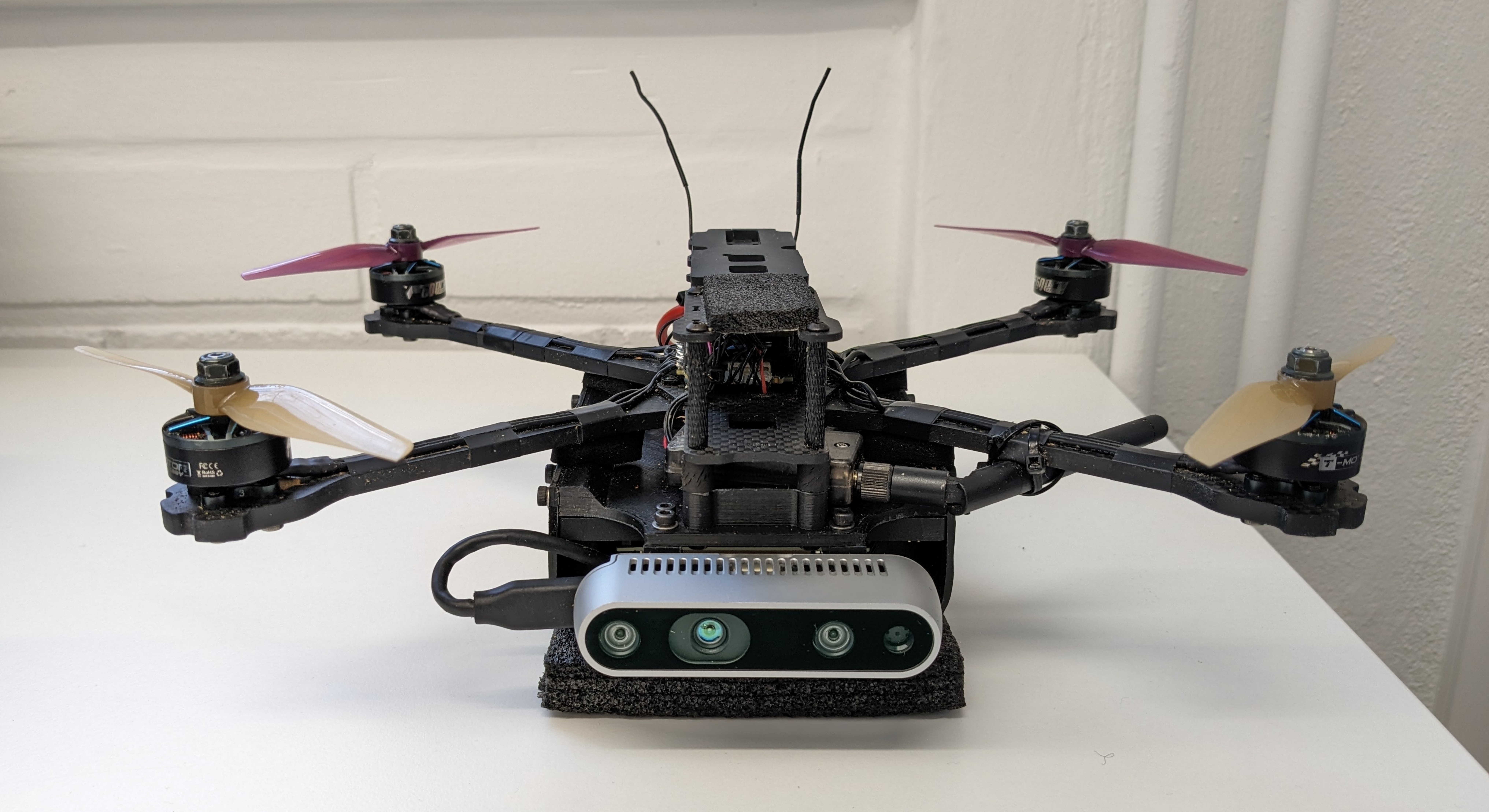}
	\caption{The drone hardware used in the flight tests with a forward-facing Realsense D435 camera. Nvidia Jetson Orin NX onboard computer is hidden in the case under the drone frame. Holybro Kakute H7 Flight controller unit is located inside the frame. Batteries (not visible in the picture) are mounted on the top of the drone frame.}
\label{fig:drone_hw}
\end{center}
\end{figure}

The computer, the camera, and the autopilot were attached to a 330 mm drone frame. The propellers had two pieces of 6 cm long blades each. The measured weight of the drone, including the motors and the onboard computer, was 791 g without batteries and 1153 g with the batteries.
An overview of drone hardware is presented in Figure \ref{fig:drone_hw}.

\subsection{Test sites}\label{sec:Test environment}

The performance of the system was evaluated in three different boreal forest areas. The flight tests of the system were performed in two test areas located in Evo, Finland (61.19\textdegree{}N, 25.11\textdegree{}E). The flight paths went through well-documented forest sample plots that have been widely used in previous studies, such as in \citet{liang2018situ, liang2019forest, wang2019field, hyyppa2020under, hyyppa2020comparison}, and are classified into three complexity categories, i.e., 'easy', 'medium', and 'difficult' based on, mainly tree density and amount of vegetation under the canopy in the plots. The third test area, a spruce forest in Palohein{\"a}, Finland (60.26\textdegree{}N, 24.92\textdegree{}E) was used only to test the loop detection performance of VINS-Fusion.

\begin{figure}[ht!]
\begin{center}
		\includegraphics[width=1.0\columnwidth]{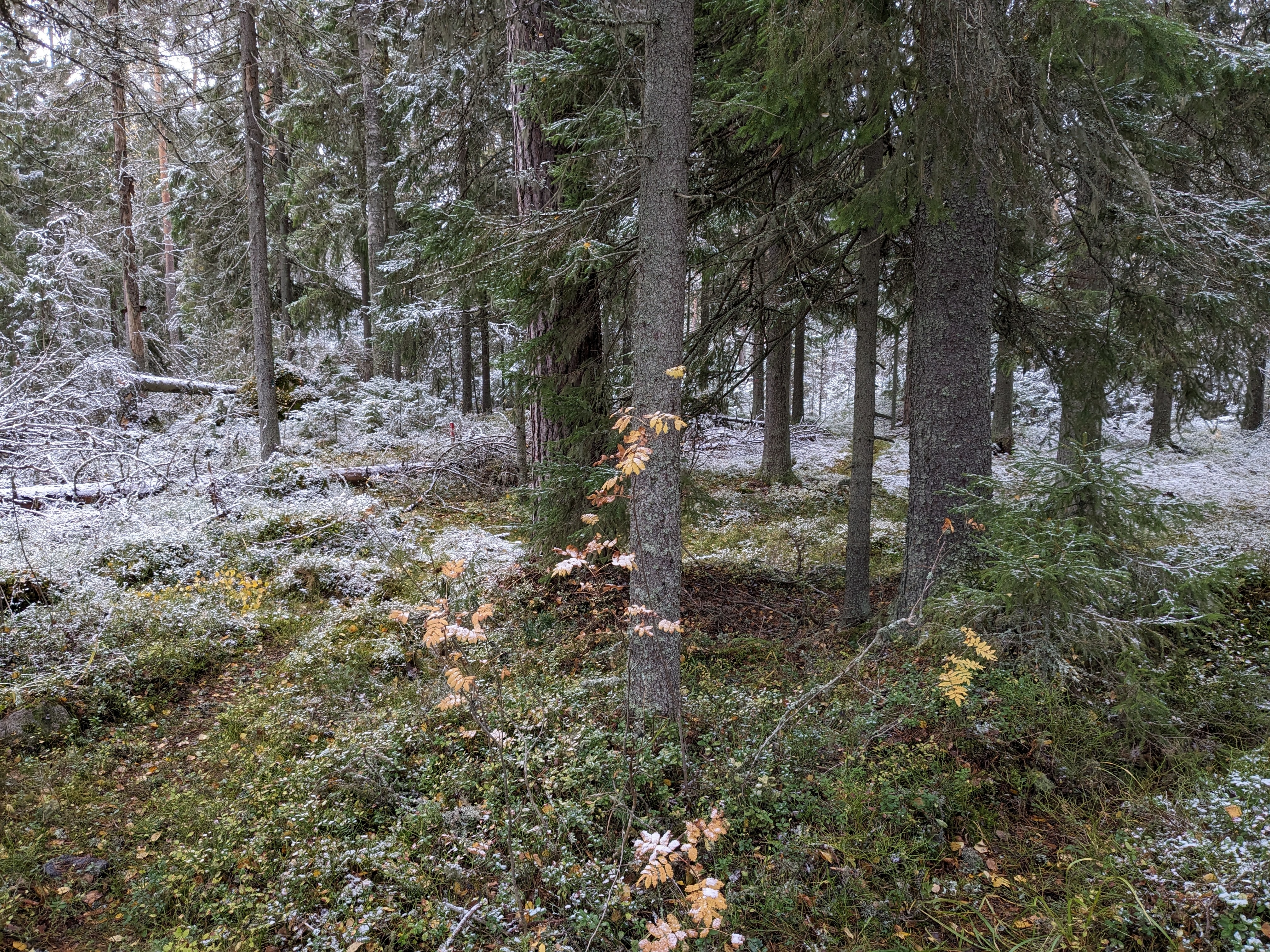}
	\caption{ View from the first takeoff location in the 'Evo-medium' test area. The takeoff location was changed after five test flights.}
\label{fig:evo_medium}
\end{center}
\end{figure}

\begin{figure}[ht!]
\begin{center}
		\includegraphics[width=1.0\columnwidth]{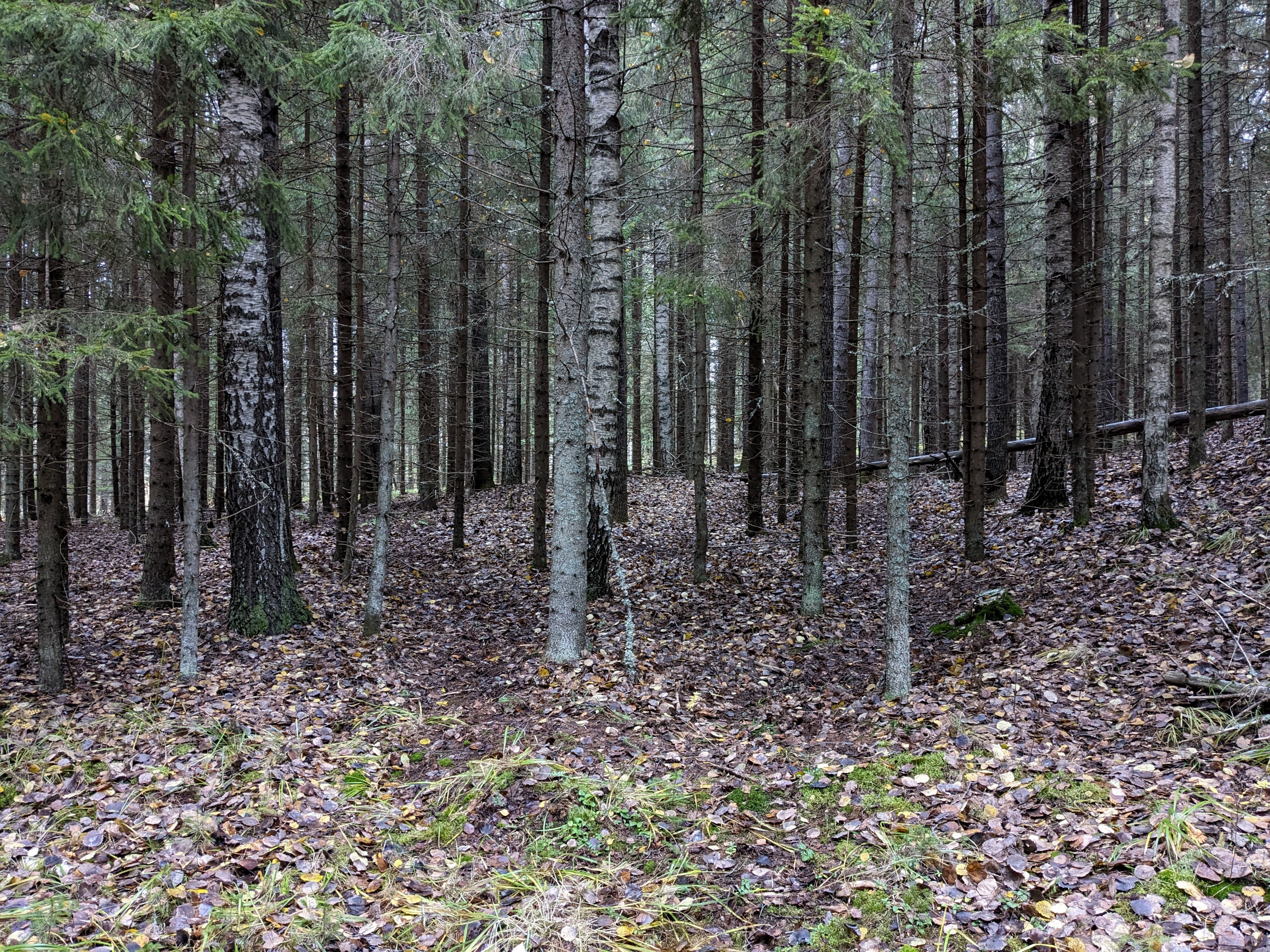}
	\caption{View from the takeoff location in the 'Evo-difficult' test forest.}
\label{fig:evo_difficult}
\end{center}
\end{figure}

The first plot in Evo has been classified as 'medium' (Figure \ref{fig:evo_medium}), and the second has been classified as 'difficult' (Figure \ref{fig:evo_difficult}). The medium-difficulty plot had a tree density of 650 trees/ha and a mean DBH of 28 cm. The most common tree species in the plot were Norway spruce (\textit{Picea abies} (L.) H.Karst.) (81.2 \%) and Scots pine (\textit{Pinus sylvestris} L.) (9.4 \%). The difficult forest had a tree density of 2000 trees/ha and a mean DBH of 17 cm. The most common tree species in the plot were Norway spruce (64.4 \%) and aspen (\textit{Populus tremula}, L.) (18.3 \%). Only trees with a DBH greater than 5 cm were included in these statistics. From now on, these two test areas are called 'Evo-medium' and 'Evo-difficult', respectively.

\begin{figure}[ht!]
\begin{center}
		\includegraphics[width=1.0\columnwidth]{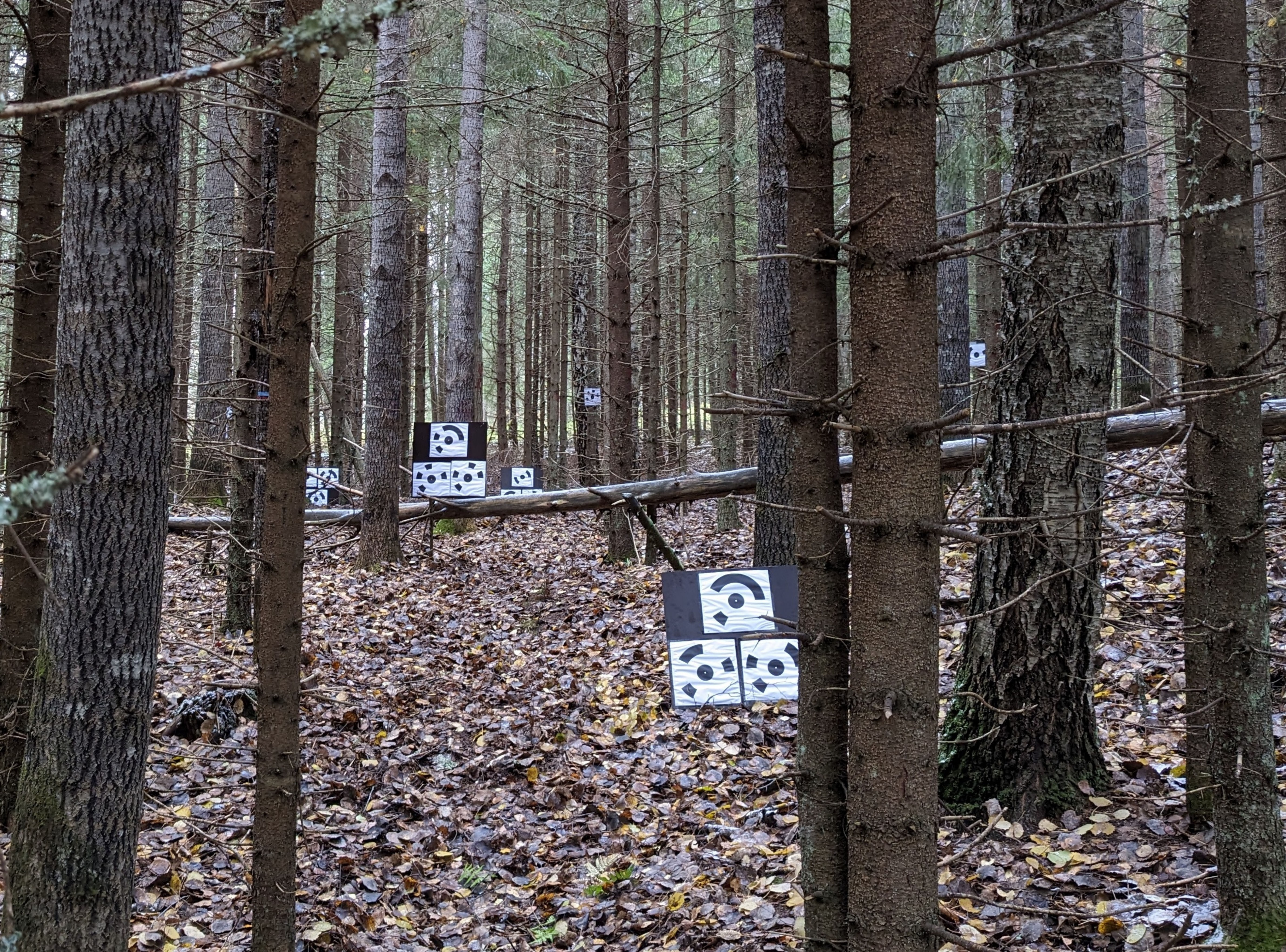}
	\caption{GCPs for reference trajectory generation in 'Evo-difficult' forest. Targets were spread widely to the test site to increase the probability that there would be targets visible in the recorded camera data regardless the path that the robot takes.}
\label{fig:evo_targets}
\end{center}
\end{figure}

For reference purposes, circular Ground Control Points (GCPs) provided by Agisoft Metashape (Agisoft LLC, St. Petersburg, Russia) were placed within the forest plots used in the flight tests. The targets had a diameter of 24.7 cm and the middle dots had a diameter of 7 cm. Relative distances between the GCPs were measured with a Leica Nova TS60 total station with a precision of approximately 2 mm \citep{TS60Datasheet}. Figure \ref{fig:evo_targets} shows the GCPs in 'Evo-difficult' during the test flights. 

GCPs were not used in the third test forest, 'Palohein{\"a}'. The experiment there aimed to test the loop detection performance of VINS-Fusion in a homogeneous forest environment without artificial targets. The data sequences were collected by walking with the drone. The estimated average tree density in the area was approximately 2000 trees/ha.

\subsection{Flight tests} \label{sec:flights}

Test flights in Evo were conducted on 18th October 2023. Multiple test flights were conducted to evaluate forest-type effects on autonomous navigation performance. In total, seven test flights were conducted in the 'Evo-medium' test area and nine test flights in the 'Evo-difficult' test area. In each test environment, either the takeoff location or the takeoff heading was altered during the test flight campaign to force the drone to follow different paths. In the 'Evo-medium' test area, the takeoff location was changed after five test flights. In the 'Evo-difficult' test area, the takeoff heading was changed after six test flights. The target flight velocity was 1 m/s in all flight tests. The goal was set 36 m forward in the first takeoff location in the 'Evo-medium' test area, 34 m forward in the second takeoff location in the 'Evo-medium' test area, and 42 m forward in the takeoff location in the 'Evo-difficult' test area. The maximum flight height was set to 2.25 m relative to the takeoff location in the 'Evo-medium' test area. For the first test flight in the 'Evo-difficult' test area, the maximum flight height was set to 2.5 m with respect to the takeoff location and was increased to 2.75 m for the rest of the flights.  

The campaign to evaluate the loop detection capability took place on 26th January 2024. Stereo images and IMU data were collected during two walks inside the forest while carrying the drone by hand. The paths of the walks were approximately 340 m long and 240 m long, each containing multiple loops where the same scene was observed multiple times. Data collection was carried out during winter, with the forest covered in snow.

\subsection{Structure-from-Motion Multi-View Stereo based reconstruction} \label{sec:Structure-from-motion}

Photogrammetric Structure-from-Motion Multi-View Stereo (SfM-MVS) processing was conducted in post-processing mode to generate 3D point clouds for DBH estimation and to generate reference trajectories for evaluating VIO performance. 

Stereo image processing was carried out using Agisoft Metashape software (Agisoft LLC, St. Petersburg, Russia) \citep{Metashape} following similar processing steps as introduced by \citet{Viljanen2018}. Every sixth stereo pair was used in the Metashape processing with a stereo baseline of 5 cm and an accuracy setting of ±0.0015 m. In the image orientation processing, the quality setting was set to high, meaning that two times downsampled images were used. The settings for the number of key points and tie points per image were 40,000 and 4,000, respectively. To enhance the quality of the post-processed reference trajectories, GCPs were used in the processing with an accuracy setting of $\pm 0.005$ m. Automatic outlier removal was performed using the gradual selection tools of the software based on re-projection error, reconstruction uncertainty, and projection accuracy. Subsequently, a bundle adjustment was carried out, involving optimization of the sparse point cloud along with the interior (IOP) and the exterior orientation parameters (EOP). Following that, a dense point cloud was calculated using the highest quality setting and mild depth filtering. Finally, the EOPs and dense point clouds were exported.

Point clouds for DBH estimation were derived using datasets from four flights conducted in the 'Evo-medium' test area, each designated as follows: I: the first flight,
II: the second flight,
III: the third flight,
IV: the fourth flight, and I-IV: combined processing of all four flights (Combined). The processing was carried out using all GCPs. Additionally, to evaluate the impacts of GCPs, point clouds were generated from the first flight trajectory processed with different GCP configurations, denoted as:
I$_{0GCP}$: no GCPs,
I$_{1GCP}$: one GCP at the beginning of the trajectory,
I$_{2GCP}$: two GCPs, one at the beginning and one at the end of the trajectory, and I$_{3GCP}$: three GCPs, positioned at the beginning, middle, and end of the trajectory.

\begin{figure}[ht!]
\begin{center}
    \includegraphics[width=1.0\columnwidth]{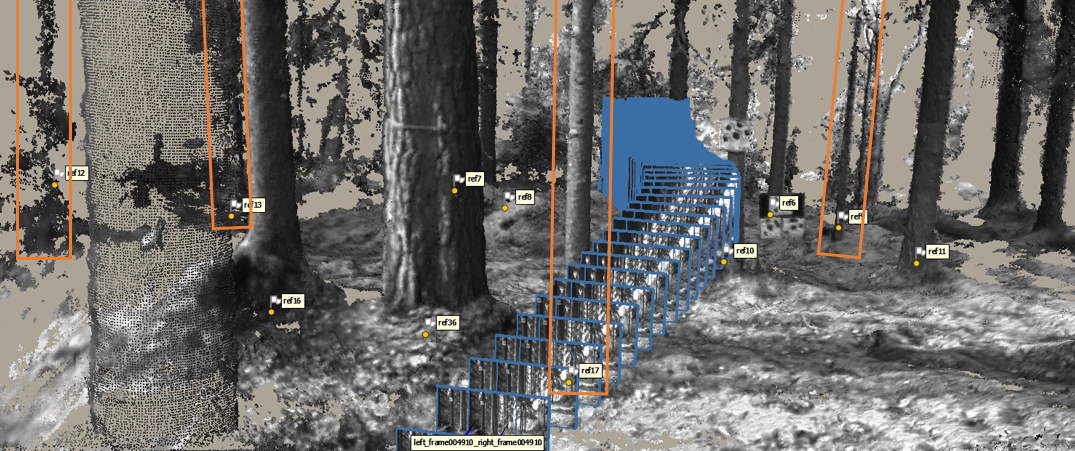}
    \caption{GCPs, stereo images and reference trees visualized during the point cloud processing in the Agisoft Metashape (Metashape Development Team, 2021). Detected reference trees are marked, while those not detected are highlighted with a orange box.}
\label{fig:gcps-and-reftrees-b}
\end{center}
\end{figure}

The experiments were conducted in dense forests, where collecting reference data using GNSS was not possible due to the forest canopy blocking the signal. Reference trajectories were generated for the flights using SfM processing. In each case, the reference trajectory was derived from the first takeoff heading and orientation in both Evo test environments. Figure \ref{fig:gcps-and-reftrees-b} visualizes the GCPs, stereo images, and reference trees during the point cloud processing in the Agisoft Metashape.

\subsection{Tree parameter estimation} \label{sec:treeParameters}

In this study, trees were detected, and their DBHs were estimated using an automatic stem diameter estimation algorithm proposed by \citet{hyyppa2020accurate} and further developed by \citet{hyyppa2020under} to be suitable for a larger variety of point clouds, including photogrammetric point clouds, and point clouds produced with 3D mobile laser scanners and TLS. The method described by \citet{hyyppa2020under} was applied to estimate the DBH from point clouds collected with the drone and a TLS system, the latter of which provided reference measurements of DBH (see details in Section \ref{sec:Performance}).

The point cloud was first segmented into 40-cm-tall (20-cm-tall for TLS) vertical segments and 5-s-long temporal segments. The temporal segmentation was done to mitigate point cloud distortions caused by potential slowly-accumulating positioning errors of the moving drone system, and thus, the temporal segmentation was not performed for the TLS data. The points potentially corresponding to the tree stems were detected by clustering the points with density-based clustering (DBSCAN, \citep{ester1996DBSCAN}) and filtered by fitting a circle to each cluster while accepting only those clusters satisfying the following heuristic criteria: the cluster contained more than 35 points and at least 80 \% of the points within 30 mm of the circular arc, the fitted circle radius was between 4 cm and 40 cm, and the points covered at least a 60\textdegree{} central angle of the fitted circle. The accepted clusters were further grouped into tree stems by applying DBSCAN for the center points of the clusters, and the growth direction of each tree was estimated with principal component analysis (PCA). The points within accepted clusters were projected into the plane perpendicular to the growth direction for the final diameter estimation. The diameters were processed into a stem curve by automatically removing clear outliers, followed by fitting a smoothing spline to the remaining diameter estimates as a function of height. Finally, the DBH estimates were calculated by interpolating the stem curve at a height of 1.3 m above the ground. For a detailed description of the method, the reader is referred to \citet{hyyppa2020accurate, hyyppa2020under}.

\subsection{Performance assessment} \label{sec:Performance}

\subsubsection{Autonomous flying performance} \label{sec:FlyingPerformance}
The performance and reliability of the autonomous flight of the prototype were evaluated based on two factors: the reliability of the obstacle avoidance and the accuracy of the VIO localization. 

The reliability of obstacle avoidance was evaluated on the basis of mission success. A mission was considered successful if the drone reached the end of the flight without major collisions. Minor touches to thin vegetation or branches were allowed if the drone was able to continue the mission after that. If the given goal was inside an obstacle, the mission was considered successful if the drone managed to fly next to the goal. In addition to obstacle avoidance, the smoothness of the trajectory was assessed. The trajectory was considered smooth if the system did not perform any emergency stops during the flight.

The accuracy of the VIO localization was evaluated by calculating the Absolute Trajectory Error (ATE), as proposed by \citet{sturm2012ate}, with respect to the post-processed SfM reference trajectories. In this study, ATE was calculated only for the position estimate, and the error in the orientation estimate was not evaluated. ATE was calculated using the trajectory evaluation toolbox by \citet{rpgToolbox}, which aligns the trajectory estimates with the reference trajectories with the method by \citet{Umeyma1991}. After aligning the trajectories, the toolbox calculates the RMSE of the individual state errors to generate a single metric error value representing the accuracy of the entire trajectory. In the mathematical form, the error can be presented as
\begin{equation}
    \text{ATE}_{\text{pos}} = \left( \frac{1}{N} \sum_{i=0}^{N-1} ||\Delta \mathbf{p}_i ||^2  \right)^\frac{1}{2}
\end{equation}
where $\Delta \mathbf{p}_i$ is the error between the estimated position and reference trajectory position in a single state.

In the first tests conducted in a simulator as described by \citep{karjalainen2023}, the VINS-Fusion estimate was found to be more accurate when the $t_d$ value was fixed to a constant, compared to when the online estimation method \citep{onlineTd} was used. For this reason, during the test flights, the $t_d$ value was fixed to the one obtained from kalibr. However, the calibration estimate is never perfect, and the prototype did not have actual hardware synchronization between the camera and the autopilot IMU. For that reason, after the test flights, the VINS-Fusion estimation was also run with online estimated $t_d$ with the camera and IMU data recorded during the flight tests. The estimates were performed with the onboard computer of the drone, and the $t_d$ estimate obtained with kalibr was used as the initial value of the online estimation.    

The loop detection capability of VINS-Fusion was evaluated in a natural forest environment in Palohein{\"a},  without the use of artificial targets that could affect the system performance. Since there were no GCPs to form the reference trajectories, only the ability to detect loops inside dense natural forests was tested and the accuracy of the loop-corrected trajectory was not evaluated. VINS-Fusion estimate with the loop detection algorithm was run for the collected data with the onboard computer of the drone, and the effect of a maximum number of features in pictures for loop detection capability was examined.

\subsubsection{DBH estimation performance}\label{sec:DBHEstimation}

The accuracy of the DBH estimation from the drone point cloud data was evaluated using TLS data as a reference. The TLS scans were collected in April 2020, and tree height, DBH, and other parameters were estimated from the dataset. The estimated error of the DBH measurement was less than 1 cm. 

Several metrics were used to evaluate the accuracy of the DBH estimation, including RMSE, Relative RMSE (RMSE \%), Bias, Relative Bias (bias \%), and Standard Deviation. 

The RMSE measures the square root of the average of the squared differences between the estimated and reference DBH values. Bias indicates the average deviation between the estimated and reference DBH values.

\begin{equation}
\text{RMSE} = \sqrt{\frac{1}{n} \sum_{i=1}^{n} \left( d_{\text{estimated},i} - d_{\text{reference},i} \right)^2 }
\end{equation}
where $n$ is the total number of observations, $d_{\text{estimated},i}$ is the estimated DBH, and $d_{\text{reference},i}$ is the reference DBH.

\begin{equation}
\text{Bias} = \frac{1}{n} \sum_{i=1}^{n} \left( d_{\text{estimated},i} - d_{\text{reference},i} \right)
\end{equation}

Relative RMSE (RMSE \%) is computed by dividing the RMSE by the mean of the reference DBH values and multiplying by 100 to express it as a percentage. Relative Bias (Bias \%) is determined by dividing the Bias by the mean of the reference DBH values and multiplying by 100.
\begin{equation}
\text{Relative RMSE (\%)} = \frac{\sqrt{\frac{1}{n} \sum_{i=1}^{n} \left( d_{\text{estimated},i} - d_{\text{reference},i} \right)^2 }}{\bar{d}_{\text{reference}}} \times 100
\end{equation}

\begin{equation}
\text{Relative Bias (\%)} = \frac{\frac{1}{n} \sum_{i=1}^{n} \left( d_{\text{estimated},i} - d_{\text{reference},i} \right)}{\bar{d}_{\text{reference}}} \times 100
\end{equation}
Additionally, standard deviation was calculated with the following equation:
\begin{equation}
\text{SD} = \sqrt{\frac{1}{n} \sum_{i=1}^{n} \left( d_{\text{estimated},i} - \bar{d}_{\text{reference}} \right)^2}
\end{equation}
where $\bar{d}_{\text{reference}}$ is the mean of the reference DBH values, and $\text{SD}$ is the standard deviation.

The completeness of stem detection was evaluated by comparing the total number of reference trees found from the drone-collected point cloud data to the total number of reference trees within the point cloud boundary.
\begin{equation}\label{eq:completeness}
c = \frac{r}{t} \times 100
\end{equation}
where $c$ represents completeness (\%), $r$ is the number of reference trees found, and $t$ is the total number of reference trees.

The impact of tree growth between the collection of drone data and reference measurements (four growth seasons) was not taken into account. Variability in tree growth, influenced by factors such as species, locations, health, and age, is acknowledged as a source of error. Considering that the tree growth was relatively constant in the test area, it formed a major component of the bias in the estimates. Additionally, some reference trees fell between the measurements and were not included in the estimations.

\section{Results}\label{sec:Results}

\subsection{Autonomous navigation performance}\label{sec:navi results}

\subsubsection{Obstacle avoidance} \label{sec:obstacle avoidance}

The task of avoiding collisions during the flight is naturally more demanding in a dense forest than in a sparse forest. Consistent with these expectations, the trajectories were smoother and the success rate was better in 'Evo-medium' than in 'Evo-difficult' test area. In 'Evo-medium', all performed test flights were successful, and five out of seven test flights were performed with smooth trajectories without any emergency stops during the flight. In 'Evo-difficult', one of nine test flights failed. None of the test flights in 'Evo-difficult' was performed with a smooth trajectory. However, despite emergency stops during the flight, the system was still able to continue the mission after stops, leading to a successful mission on eight of the nine test flights. 

The reason for emergency stops in both environments was the late detection of thin and dry spruce branches. In the failed test flight in 'Evo-difficult', the reason was also a collision with a similar thin low branch. 

\begin{figure*}[h!]
\begin{subfigure}{\textwidth}
\begin{subfigure}{0.5\textwidth}
\includegraphics[width=0.95\linewidth]{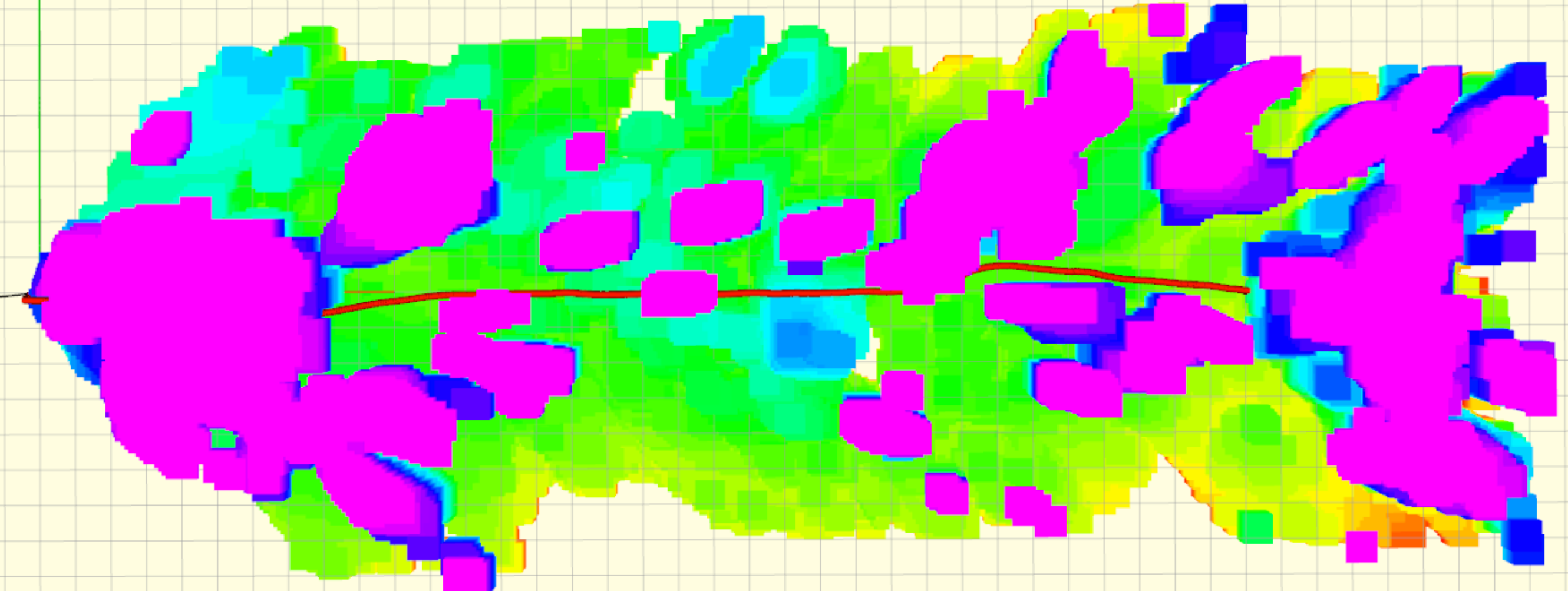} 
\centering
\label{fig:subim1}
\end{subfigure}
\begin{subfigure}{0.5\textwidth}
\includegraphics[width=0.95\linewidth]{2023-10-18-13-14-17_ccompare.pdf}
\centering
\label{fig:subim2}
\end{subfigure}
\caption{'Evo-medium'}
\end{subfigure}
\medskip

\begin{subfigure}{\textwidth}
\begin{subfigure}{0.5\textwidth}
\includegraphics[width=0.95\linewidth]{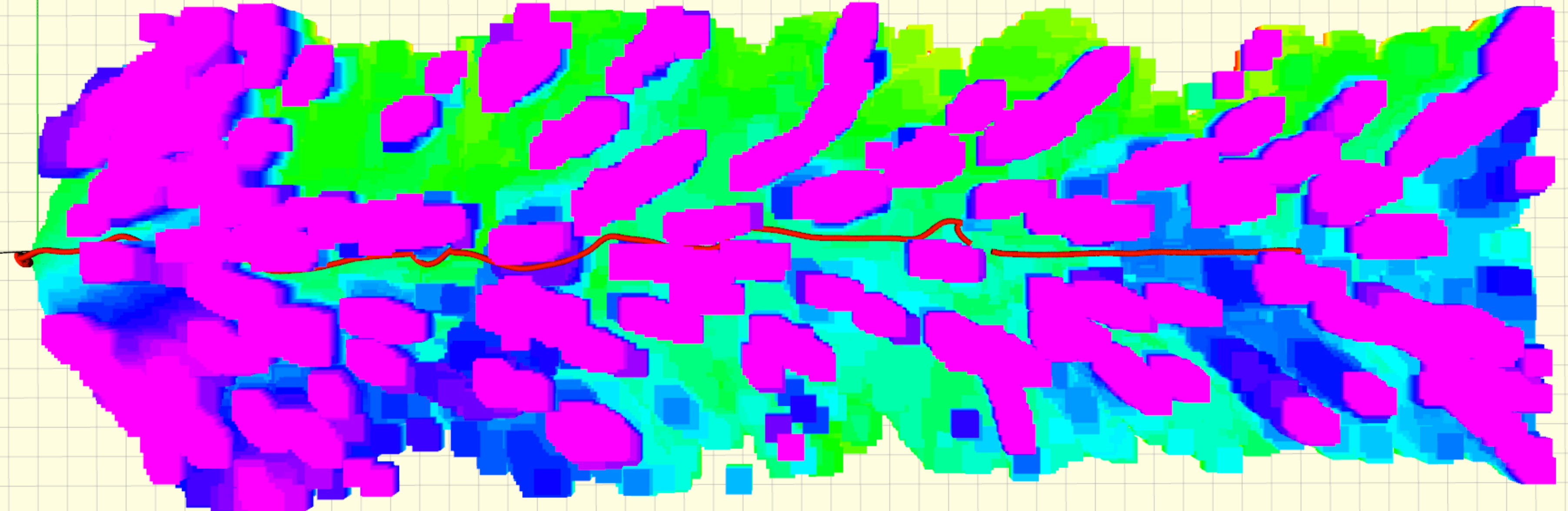} 
\centering
\label{fig:subim3}
\end{subfigure}
\begin{subfigure}{0.5\textwidth}
\includegraphics[width=0.95\linewidth]{2023-10-18-15-47-11_ccompare.pdf}
\centering
\label{fig:subim4}
\end{subfigure}
\caption{'Evo-difficult'}
\end{subfigure}

\caption{Examples of drones' own navigation grid maps with performed trajectories (left) and obtained point clouds (right) in (a) 'Evo-medium' and (b) 'Evo-difficult' }
\label{fig:paths}
\end{figure*}

The test flight success rates in both test forests are summarized in Table \ref{tab:success}. Figure \ref{fig:paths} presents examples of performed trajectories and obtained point clouds in both test forests.

\begin{table*}[!h]
\caption{Test flight success in the test forests}
\begin{center}
\resizebox{\textwidth}{!}{%
  \begin{tabular}{cccccc}
  \hline 
    \textbf{Test areas} & \textbf{Forest type} & \vtop{\hbox{\strut \textbf{Density}}\hbox{\strut \textbf{(trees/ha)}}} & \vtop{\hbox{\strut \textbf{Flight}}\hbox{\strut \textbf{lengths (m)}}}  & \vtop{\hbox{\strut \textbf{Successful}}\hbox{\strut \textbf{flights}}}  & \vtop{\hbox{\strut \textbf{Smooth}}\hbox{\strut \textbf{trajectories}}} \\
    \hline 
    'Evo-medium' & Old Spruce forest & 650 & 34 - 36 & 7/7 & 5/7\\
    'Evo-difficult' & Young Spruce forest & 2000 & 42 & 8/9 & 0/9 \\
    \hline
\end{tabular} 
}
\label{tab:success}
\end{center}
\end{table*}

\subsubsection{VIO} \label{sec:VIO}

The trajectory estimate error against the reference trajectory was calculated for the first five flights in 'Evo-medium' and for three flights in the difficult forest. The flights from the second takeoff location in 'Evo-medium' and from the second takeoff orientation in 'Evo-difficult' were not used in the VIO estimation since GCPs used in the reference trajectory processing were positioned along the first flight trajectories in both environments. In 'Evo-difficult', one of the flights from the first takeoff orientation was performed without GCPs, leading to a non-comparable reference trajectory, one of the flights had a data recording error during a flight, and in one flight, Metashape failed to retrieve the reference path. 

\begin{figure}[h!]
\begin{subfigure}{\linewidth}
\includegraphics[width=0.475\textwidth]{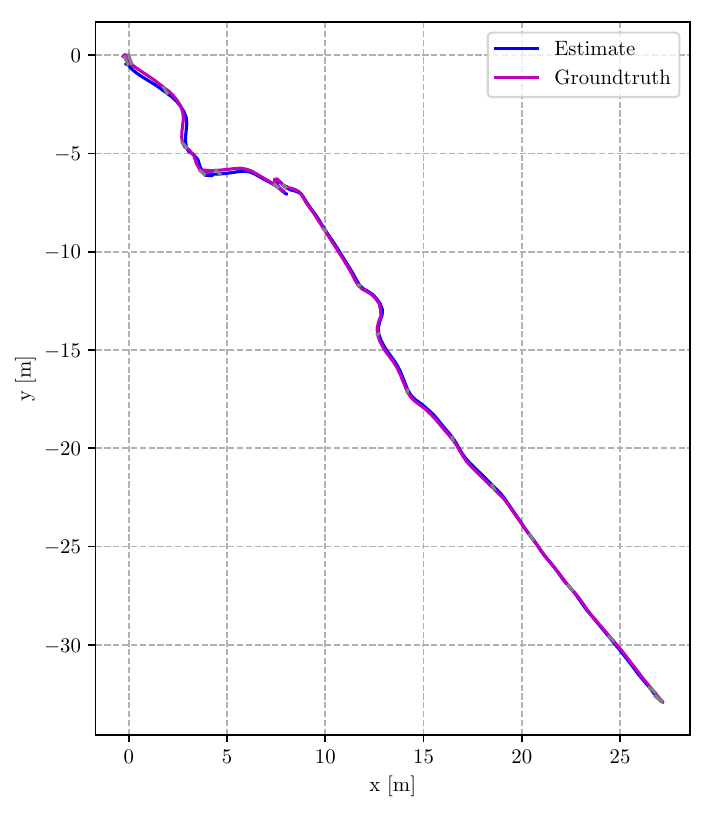} 
\centering
\label{fig:subim11}
\caption{Top view.}
\end{subfigure}
\medskip

\begin{subfigure}{\linewidth}
\includegraphics[width=\linewidth]{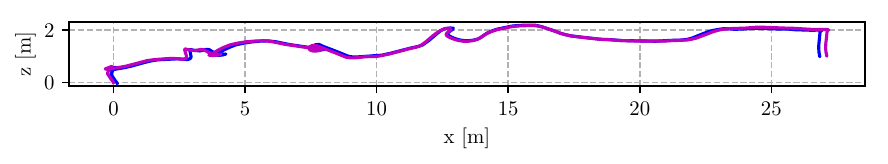} 
\centering
\label{fig:subim33}
\caption{Side view.}
\end{subfigure}
\caption{ Example of a VINS-Fusion trajectory estimate and a reference trajectory obtained with Agisoft Metashape in the 'Evo-difficult' forest from a) top and b) side.}
\label{fig:vins_traj}
\end{figure}

Figure \ref{fig:vins_traj} shows an example of the VINS-Fusion trajectory estimate compared to the reference in 'Evo-difficult', while Table \ref{tab:VIO_errors} summarizes the VINS-Fusion estimation errors across all flights. The average $\text{ATE}_{\text{pos}}$ was 0.50 m in 'Evo-medium' and 0.34 m in 'Evo-difficult' with standard deviations of 0.06 m and 0.07 m, respectively, without online $t_d$ estimation. 
The VINS-Fusion estimates with online $t_d$ estimation were computed post-flight for the same trajectories, yielding an average $\text{ATE}_{\text{pos}}$ of 0.47 m in 'Evo-medium' and 0.33 m in 'Evo-difficult' with standard deviations of 0.02 m and 0.04 m, respectively.

\begin{table*}[!h]
\caption{VINS-Fusion estimation Errors in all test flights}
\begin{center}
\resizebox{\textwidth}{!}{%
  \begin{tabular}{lccc}
  \hline 
    \textbf{Test area} & \textbf{Flight No.} & \textbf{$\text{ATE}_{\text{pos}}$ (constant $t_d$)}  & \textbf{$\text{ATE}_{\text{pos}}$ (online estimated $t_d$)}  \\
    \hline 
    'Evo-medium' & 1 & 0.5567 & 0.4816  \\
    'Evo-medium' & 2 & 0.5599 & 0.4790  \\
    'Evo-medium' & 3 & 0.4538 & 0.4769  \\
    'Evo-medium' & 4 & 0.4945 & 0.4722  \\
    'Evo-medium' & 5 & 0.4313 & 0.4400  \\
    'Evo-difficult' & 1 & 0.3787 & 0.3340  \\
    'Evo-difficult' & 2 & 0.3853 & 0.3727   \\
    'Evo-difficult' & 3 & 0.2680 & 0.2962\\ 
    \hline
\end{tabular}  
}
\label{tab:VIO_errors}
\end{center}
\end{table*}

The small standard deviations between the $\text{ATE}_{\text{pos}}$ estimates indicated that the error between the VINS-Fusion trajectories and the reference trajectories was similar across all flights. In each case, VINS-Fusion accurately captured the trajectory shape but systematically underestimated the total flight distance.

\begin{figure}[ht!]
\begin{center}
		\includegraphics[width=1.0\columnwidth]{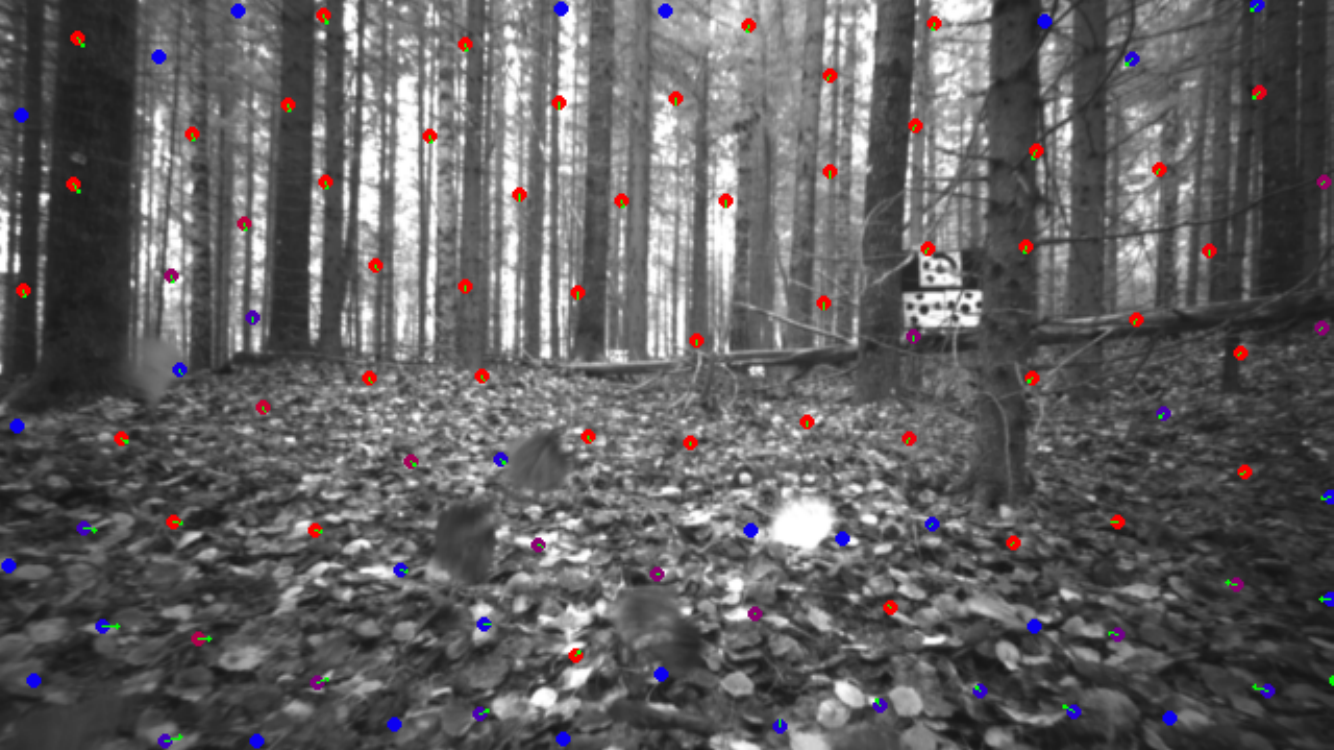}
	\caption{Example of VINS-Fusion tracking image visualizing the tracked features. The tracking was not disturbed by the dry leaves blown by the air flow from the propellers of the drone. The coloring of the feature markers depends on the time how long the feature has been tracked. Blue dots symbolize new features and red features have been tracked longer. }
\label{fig:leaves}
\end{center}
\end{figure}

Figure \ref{fig:leaves} shows an example of a VINS-Fusion tracking image during a test flight in 'Evo-difficult'. The algorithm detects stable features to track from all parts of the picture. The image also shows flying dry leaves blown up by the airflow from the propellers of the drone. The moving leaves did not cause visible disturbances to the tracking of the pose.

\begin{figure}[h!]
\begin{subfigure}{\linewidth}
\includegraphics[width=\linewidth]{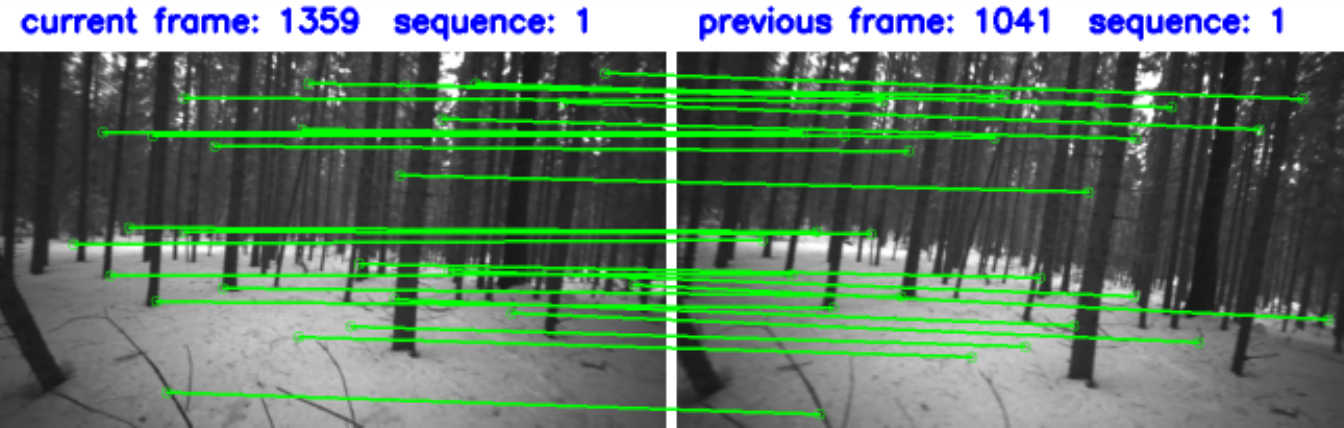} 
\centering
\label{fig:short_loop}
\caption{Shorter path.}
\end{subfigure}
\medskip

\begin{subfigure}{\linewidth}
\includegraphics[width=\linewidth]{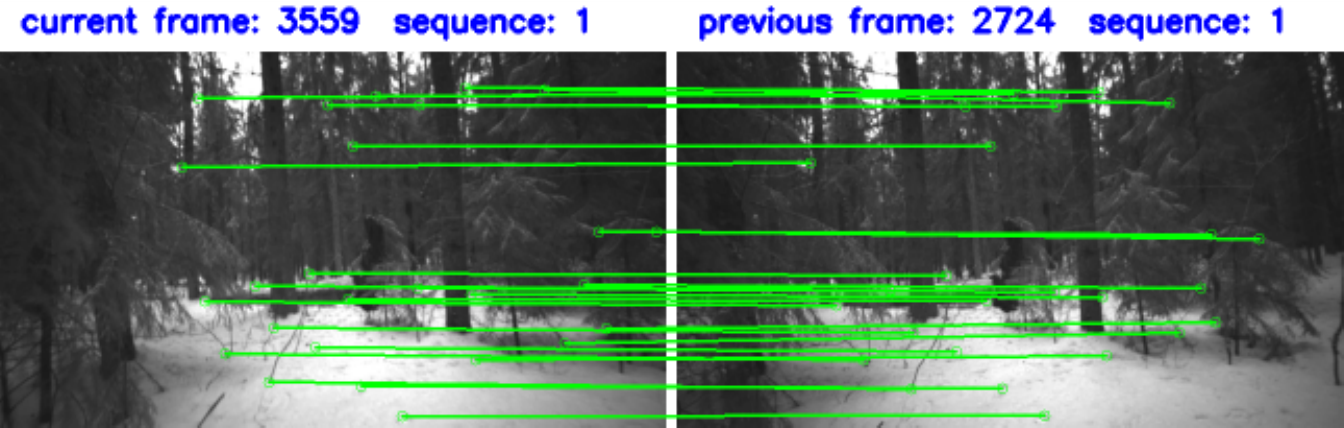} 
\centering
\label{fig:long_loop}
\caption{Longer path.}
\end{subfigure}
\caption{Example of image pairs where VINS-Fusion detects a scene to already visited one and performs a loop closure. }
\label{fig:loop_closure}
\end{figure}

In the loop closure tests, VINS-fusion detected a loop closure between 20 image pairs in the shorter path and four image pairs in the longer path with the same feature tracking parameters (Table \ref{tab:VINS-parameters}) that were used in the flight tests. Figure \ref{fig:loop_closure} shows examples of pairs of detected loop closure images from both paths. The number of detected loops can be increased by increasing the number of features per image in the VINS-Fusion configuration. By reducing the minimum distance between features from 40 to 35 pixels, VINS-Fusion detected a loop closure between 52 image pairs in the shorter path and 23 image pairs in the longer path. By decreasing the minimum distance between features to 30 pixels, VINS-Fusion detected a loop closure between 82 image pairs in the shorter path and 63 image pairs in the longer path. However, increasing the number of features also increases the computational burden of the VIO, and the effects on the real-time performance during flying were not tested in this study.

\subsection{Forest reconstruction and analysis} \label{sec:analysis results}
\subsubsection{Photogrammetric processing} \label{sec:photogrammetricproc}
Results of the photogrammetric processing are given in Table \ref{tab:stats1_pointcloud}. In the image alignment process, all images were aligned in each dataset, which indicated successful processing. Re-projection errors varied between 0.29 and 0.39 pixels. The estimated GSDs (Ground Sampling Distance) ranged between 4.08 and 6.17 mm at ground level. Distances from the trees varied between 0.55 and 7.94 m, resulting in GSDs of 0.85-12.34 mm at the objects of interest. Point clouds from datasets II-IV had similar point densities of 2.63-2.84 points/cm\textsuperscript{2}, while dataset I provided a slightly higher point density of 3.58 points/cm\textsuperscript{2}.

\begin{table*}[!htbp]
\caption{Photogrammetric processing results. Img. count: Number images/Number of aligned images; Number of GCPs; Completeness \% (Number of trees found/Number of total trees); Reprojection error (pix); Number of tie points (in millions); Point density (points/cm\textsuperscript{2}) in 'Evo-medium' test site. I: Trajectory 1, II: Trajectory 2, III: Trajectory 3, IV: Trajectory 4, Combined: Trajectories 1-4 merged, and for Trajectory 1 I$_{0GCP}$: no GCPs, I$_{1GCP}$: 1 GCP, I$_{2GCP}$: 2 GCPs, I$_{3GCP}$: 3 GCPs} 
\centering
\resizebox{\textwidth}{!}{%
\begin{tabular}{@{}ccccccc@{}}
\toprule
\textbf{Dataset} & \textbf{Img. count} & \textbf{N GCPs} & \textbf{Completeness (\%)} & \textbf{Reproj. error (pix)} & \textbf{Num. tie points (millions)} & \textbf{Point density (points/cm\textsuperscript{2})} \\ \midrule
I & 650/650 & 15 & 79.31 (23/29) & 0.39 & 197132 & 3.58 \\
II & 648/648 & 15 & 62.07 (18/29) & 0.33 & 151889 & 2.84 \\
III & 619/619 & 11 & 61.29 (19/31) & 0.33 & 151083 & 2.63 \\
IV & 597/597 & 15 & 62.50 (20/32) & 0.33 & 141716 & 2.66 \\
Combined & 2514/2514 & 15 & 33.33 (12/36) & 0.29 & 1547828 & 8.53 \\
I$_{0GCP}$ & 650/650 & 0 & 72.41 (21/29) & 0.39 & 197132 & 3.54 \\
I$_{1GCP}$ & 650/650 & 1 & 68.97 (20/29) & 0.39 & 197132 & 3.54 \\
I$_{2GCP}$ & 650/650 & 2 & 75.86 (22/29) & 0.39 & 197132 & 3.54 \\
I$_{3GCP}$ & 650/650 & 3 & 79.31 (23/29) & 0.39 & 197132 & 3.54 \\ \bottomrule
\end{tabular}%
}

\label{tab:stats1_pointcloud}
\end{table*}

Results of accuracy estimation using Check Points (CPs) are shown in Table \ref{tab:acc_cps}. 
The cases without GCPs or with one GCP at the beginning of the trajectory provided similar 2D and 3D RMSEs of 11.36-11.72 cm and 11.37-11.73 cm, respectively. When two GCPs were used, the 2D and 3D RMSEs improved to 5.26 cm and 5.48 cm, respectively. Adding an additional GCP resulted in a significant improvement in the 2D and 3D RMSE, reducing them to 1.41 cm and 1.48 cm, respectively.

\begin{table*}[ht!]
\caption{Accuracy estimation (RMSE) using Check Points (CPs) and Camera Location in 'Evo-medium' test site, for trajectory 1: I$_{0GCP}$: no GCPs, I$_{1GCP}$: 1 GCP, I$_{2GCP}$: 2 GCPs, I$_{3GCP}$: 3 GCPs}
\begin{center}
\resizebox{\textwidth}{!}{%
  \begin{tabular}{lccccccccccc}
    \hline 
    \multicolumn{1}{c}{\textbf{Dataset}} & \multicolumn{6}{c}{\textbf{Check Points (CPs) RMSE (cm)}} & \multicolumn{5}{c}{\textbf{Camera location RMSE (cm)}} \\
    & \textbf{N CPs} & \textbf{X} & \textbf{Y} & \textbf{Z} & \textbf{2D} & \textbf{3D} & \textbf{X$_{0}$} & \textbf{Y$_{0}$} & \textbf{Z$_{0}$} & \textbf{2D$_{0}$} & \textbf{3D$_{0}$}\\
    \hline
    I$_{0GCP}$ & 15 & 6.99 & 8.95 & 0.65 & 11.36 & 11.37 & 7.06 & 9.07 & 0.30 & 11.49 & 11.50\\
    I$_{1GCP}$ & 14 & 7.19 & 9.25 & 0.56 & 11.72 & 11.73 & 7.03 & 9.06 & 0.28 & 11.47 & 11.47\\
   I$_{2GCP}$ & 13 & 5.01 & 1.61 & 1.52 & 5.26 & 5.48 & 7.03 & 9.06 & 0.28 & 11.47 & 11.47\\
   I$_{3GCP}$ & 12 & 1.19 & 0.75 & 0.45 & 1.41 & 1.48 & 1.46 & 0.70 & 0.56 & 1.62 & 1.71\\
    \hline
\end{tabular} 
}
\label{tab:acc_cps}
\end{center}
\end{table*}

\subsubsection{Tree detection} \label{sec:TreeDetection}
\begin{figure*}[!ht]
    \centering
    \begin{minipage}[b]{0.45\linewidth}
        \centering
        \includegraphics[width=\linewidth]{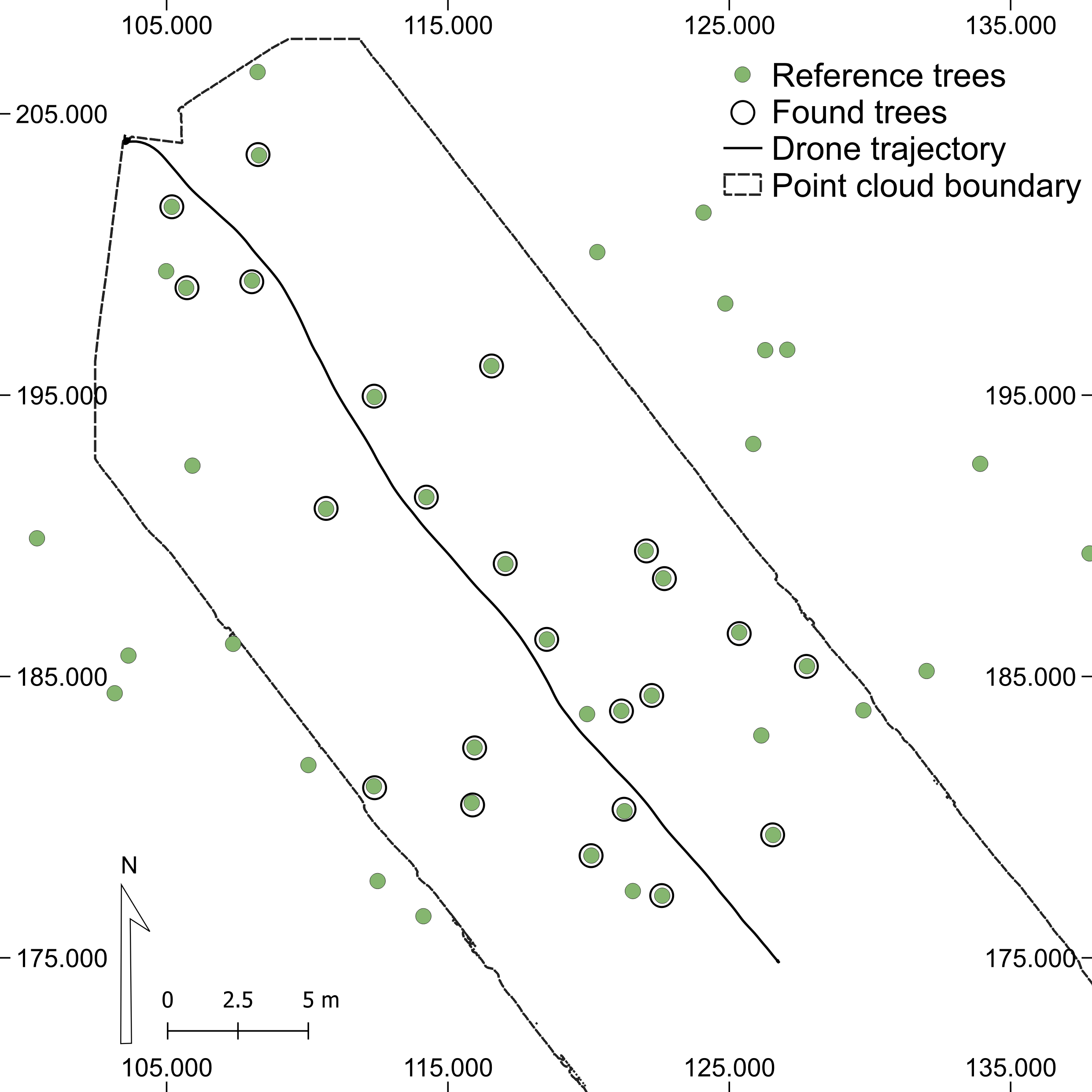}
        \caption*{(a)}
    \end{minipage}
    \hfill
    \begin{minipage}[b]{0.45\linewidth}
        \centering
        \includegraphics[width=\linewidth]{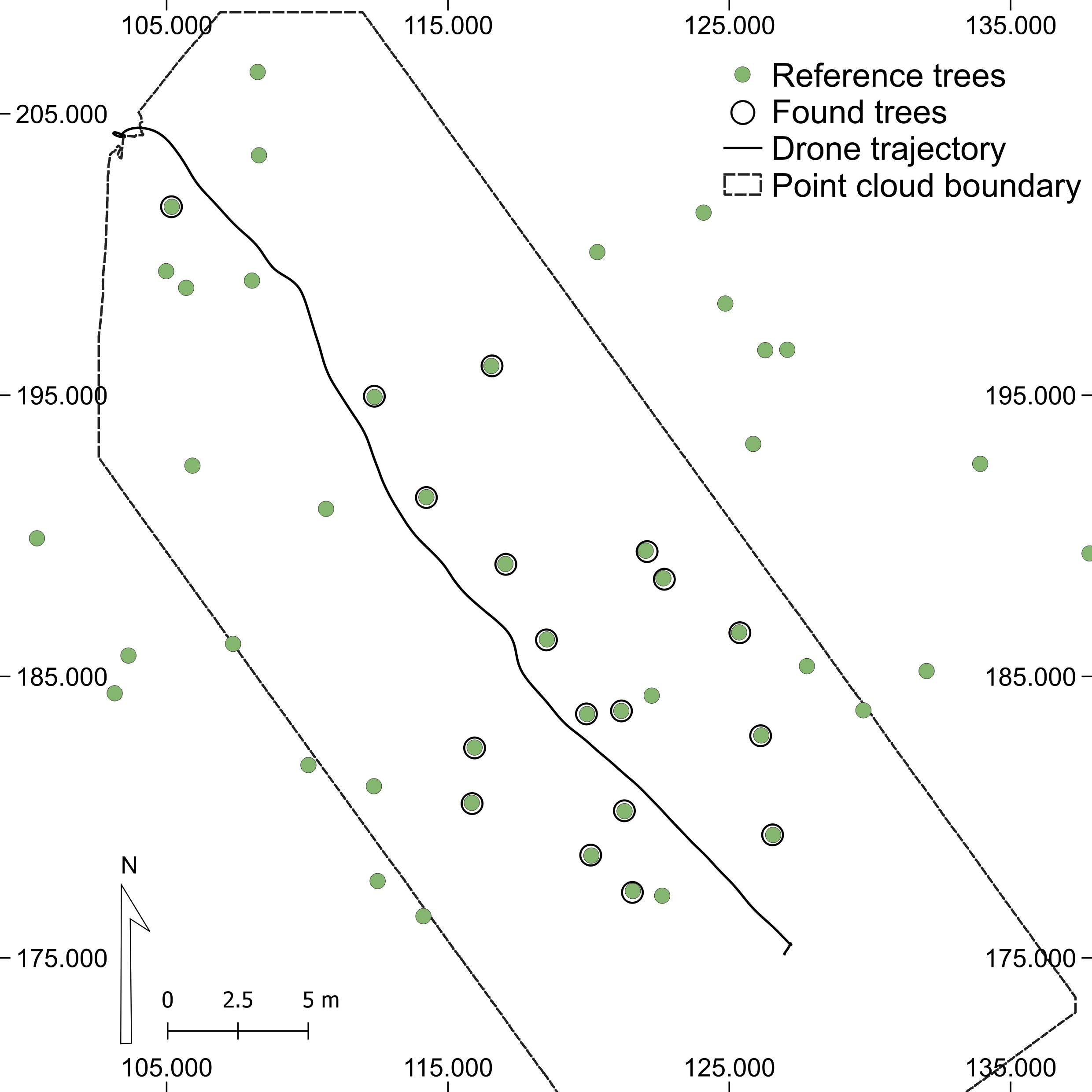}
        \caption*{(b)}
    \end{minipage}
    
    \vspace{1cm}
    
    \begin{minipage}[b]{0.45\linewidth}
        \centering
        \includegraphics[width=\linewidth]{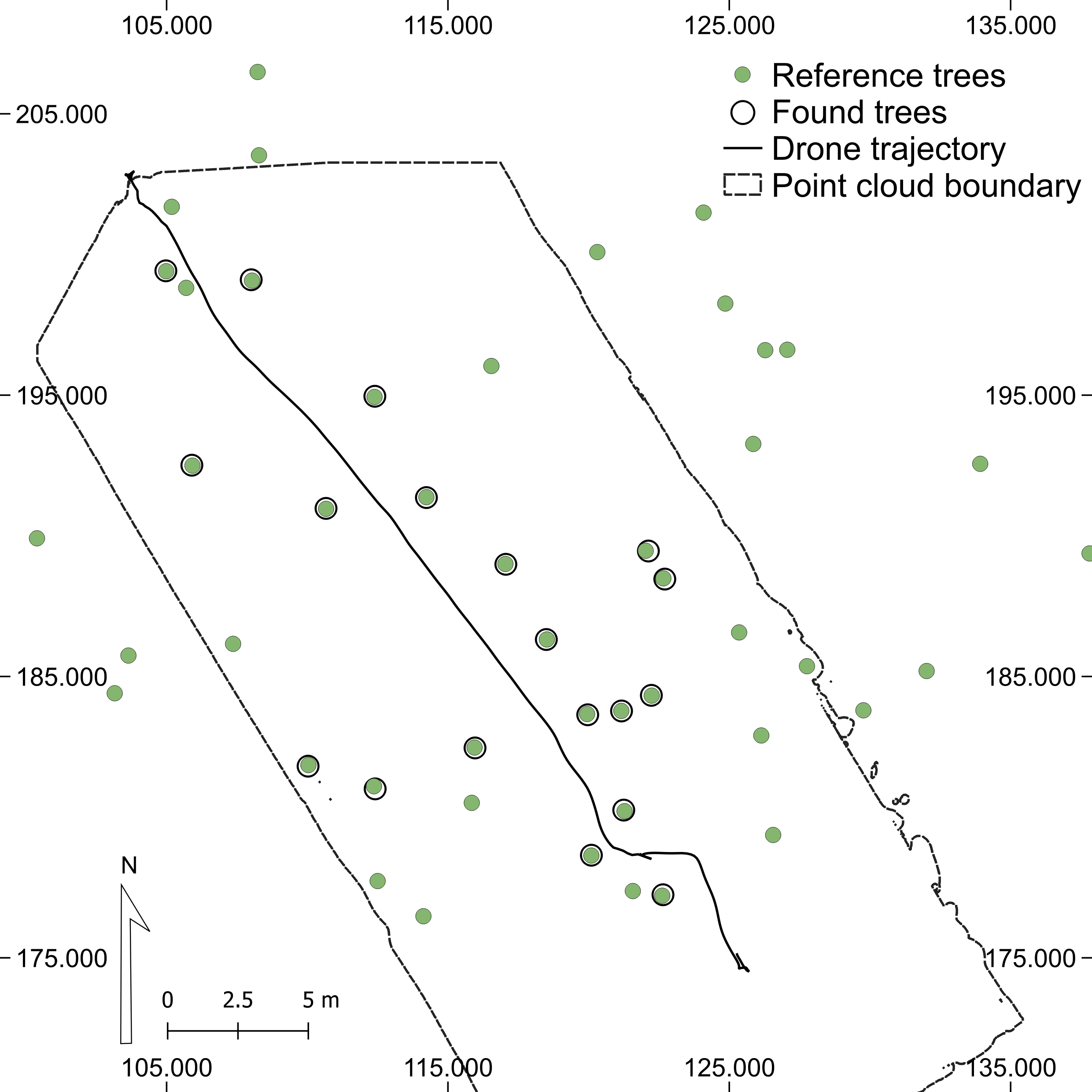}
        \caption*{(c)}
    \end{minipage}
    \hfill
    \begin{minipage}[b]{0.45\linewidth}
        \centering
        \includegraphics[width=\linewidth]{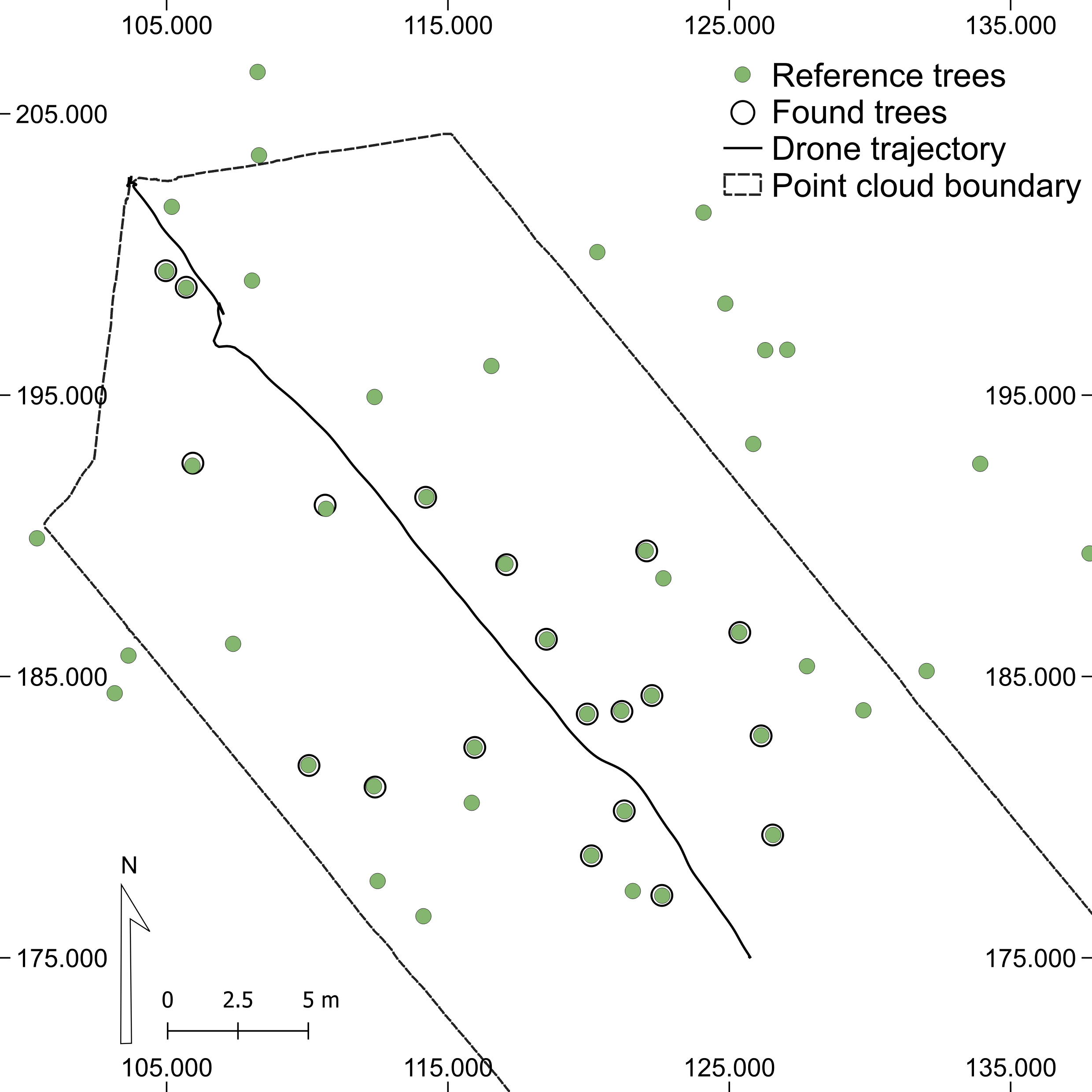}
        \caption*{(d)}
    \end{minipage}  
    \caption{Post-processed trajectories, reference trees and found trees in Dataset (a) I (b) II (c) III (d) IV}
\label{fig:trajectories-foundtrees}
\end{figure*}

Figure \ref{fig:trajectories-foundtrees} shows post-processed trajectories for dataset I-IV, as well as reference and detected trees and the boundaries of the point clouds. I and II followed almost the same trajectory, while III and IV had similar trajectories. The dataset I trajectory was the smoothest, and it did not include any sharp turns. The other three trajectories had one or more sharper turns. For datasets I, II, III, and IV, the number of reference trees inside the point cloud boundary was 29, 29, 31, and 32, of which 23, 18, 19, and 20 were found from the drone-collected point clouds, respectively. The resulting completeness values were 79.31 \%, 62.07 \%, 61.29 \% and 62.5 \%, respectively. The total number of different trees found was 29 out of 33 (87,88 \%) from dataset I-IV. However, when all flights were processed simultaneously (combined), that resulted in only 33.33 \% completeness, where only 12 of 36 trees were found. When utilizing 0, 1, 2, and 3 GCPs in the dataset I, completeness values were 72.41 \%, 68.97 \%, 75.86 \%, and 79.31 \%, respectively (Table \ref{tab:DBH_estimation_stats}).

In dataset I, the stem detection algorithm failed to detect one large tree and two small trees located near the flight trajectory, despite these trees being well reconstructed in the point cloud. Similarly, the number of trees well reconstructed but not detected by the algorithm was 4, 5, and 5, in datasets II, III, and IV, respectively. The three extra trees that were not identified in dataset I were situated further from the trajectory and exhibited poor reconstruction quality in the point cloud. The datasets II through IV each contained seven poorly reconstructed trees. Many of these trees were located at a greater distance from the trajectory, while others were obscured from the view of the camera by intervening trees.

\subsubsection{DBH estimation} \label{sec:DBHestimation}

RMSE and bias of the DBH estimates of datasets I-IV were similar, ranging from 3.33 to 3.86 cm (10.69 \%-12.98 \%) and -0.67 to 1.40 (-2.11 \%-4.71 \%), respectively (Table \ref{tab:DBH_estimation_stats}). Processing four flights simultaneously (Combined) provided only marginally better results with an RMSE of 2.91 cm (9.14 \%) and a bias of 1.51 cm (4.74 \%). 

When considering the impact of the number of GCPs, the utilization of 3, 2, 1, and 0 GCPs in photogrammetric processing yielded slightly poorer results compared to those obtained with all GCPs. The RMSE and bias values ranged from 3.93 to 4.37 cm (13.68 \%-14.66 \%) and 1.12 to 2.13 cm (3.89 \%-7.55 \%), respectively (Table \ref{tab:DBH_estimation_stats}). 

Tree size had an impact on the accuracy of DBH estimation (Table \ref{tab:DBH_estimation_stats}). For small and medium-sized trees with a DBH smaller than 30 cm, DBH estimations achieved good performance in all datasets, with RMSE and bias varying from 1.16 to 2.75 cm (5.74 \%-12.47 \%) and -0.01 to 0.62 cm (-0.06 \%-3.14 \%), respectively. Significantly poorer performance was obtained for trees with a DBH greater than 30 cm, with RMSE and bias ranging from 4.04 to 11.08 cm (8.99 \%-25.44 \%) and -0.66 to 2.08 cm (-1.53 \%-5.65 \%), respectively.

Analysis of the impact of the camera distance from trees did not reveal a significant correlation between tree distance and DBH estimation errors.

\begin{table*}[!ht]
\caption{Accuracy of the stem diameter breast height (DBH) in 'Evo-medium' test site. Dataset is split in 3 categories: All, DBH $<$ 30 cm and DBH $>$ 30 cm trees. Datasets: I: Flight 1, II: Flight 2, III: Flight 3, IV: Flight 4, Combined: Flights I-IV merged, and for Dataset I:  I$_{0GCP}$: with no GCPs,  I$_{1GCP}$: with 1 GCP,  I$_{2GCP}$: with 2 GCPs,  I$_{3GCP}$ with 3 GCPs.}
\begin{center}
\resizebox{\textwidth}{!}{%
  \begin{tabular}{ccccccc}
  \hline 
    \textbf{Dataset} & \vtop{\hbox{\strut \textbf{Number of trees}}\hbox{\strut \textbf{All}}\hbox{\strut \textbf{$<$30 cm}}\hbox{\strut \textbf{$>$30 cm}}} & \vtop{\hbox{\strut \textbf{RMSE (cm)}}\hbox{\strut \textbf{All}}\hbox{\strut \textbf{$<$30 cm}}\hbox{\strut \textbf{$>$30 cm}}} & \vtop{\hbox{\strut \textbf{RMSE \%}}\hbox{\strut \textbf{All}}\hbox{\strut \textbf{$<$30 cm}}\hbox{\strut \textbf{$>$30 cm}}} & \vtop{\hbox{\strut \textbf{Bias (cm)}}\hbox{\strut \textbf{All}}\hbox{\strut \textbf{$<$30 cm}}\hbox{\strut \textbf{$>$30 cm}}} & \vtop{\hbox{\strut \textbf{Bias \%}}\hbox{\strut \textbf{All}}\hbox{\strut \textbf{$<$30 cm}}\hbox{\strut \textbf{$>$30 cm}}} & \vtop{\hbox{\strut \textbf{Standard Deviation}}\hbox{\strut \textbf{All}}\hbox{\strut \textbf{$<$30 cm}}\hbox{\strut \textbf{$>$30 cm}}}\\
    \hline 
    I & \vtop{\hbox{\strut 23}\hbox{\strut 14}\hbox{\strut 9}} & \vtop{\hbox{\strut 3.86}\hbox{\strut 2.56}\hbox{\strut 5.29}} & \vtop{\hbox{\strut 12.98}\hbox{\strut 12.47}\hbox{\strut 11.98}} & \vtop{\hbox{\strut 1.40}\hbox{\strut 0.17}\hbox{\strut 1.23}} & \vtop{\hbox{\strut 4.71}\hbox{\strut 0.85}\hbox{\strut 2.78}} & \vtop{\hbox{\strut 3.60}\hbox{\strut 2.54}\hbox{\strut 4.26}}\\
    II & \vtop{\hbox{\strut 17}\hbox{\strut 12}\hbox{\strut 5}} & \vtop{\hbox{\strut 3.33}\hbox{\strut 2.16}\hbox{\strut 5.14}} & \vtop{\hbox{\strut 12.79}\hbox{\strut 11.36}\hbox{\strut 12.01}} & \vtop{\hbox{\strut 1.01}\hbox{\strut 0.31}\hbox{\strut 0.70}} & \vtop{\hbox{\strut 3.87}\hbox{\strut 1.63}\hbox{\strut 1.63}} & \vtop{\hbox{\strut 3.17}\hbox{\strut 2.12}\hbox{\strut 4.56}}\\
    III & \vtop{\hbox{\strut 19}\hbox{\strut 9}\hbox{\strut 10}} & \vtop{\hbox{\strut 3.97}\hbox{\strut 1.16}\hbox{\strut 5.36}} & \vtop{\hbox{\strut 12.05}\hbox{\strut 5.74}\hbox{\strut 12.07}} & \vtop{\hbox{\strut 0.97}\hbox{\strut 0.13}\hbox{\strut 0.84}} & \vtop{\hbox{\strut 2.95}\hbox{\strut 0.64}\hbox{\strut 1.89}} & \vtop{\hbox{\strut 3.85}\hbox{\strut 1.13}\hbox{\strut 5.12}}\\
    IV & \vtop{\hbox{\strut 20}\hbox{\strut 10}\hbox{\strut 10}} & \vtop{\hbox{\strut 3.40}\hbox{\strut 1.28}\hbox{\strut 4.63}} & \vtop{\hbox{\strut 10.69}\hbox{\strut 6.21}\hbox{\strut 10.77}} & \vtop{\hbox{\strut -0.67}\hbox{\strut -0.01}\hbox{\strut -0.66}} & \vtop{\hbox{\strut -2.11}\hbox{\strut -0.06}\hbox{\strut -1.53}} & \vtop{\hbox{\strut 3.33}\hbox{\strut 1.28}\hbox{\strut 4.44}}\\
    Combined & \vtop{\hbox{\strut 12}\hbox{\strut 7}\hbox{\strut 6}} & \vtop{\hbox{\strut 2.91}\hbox{\strut 1.41}\hbox{\strut 4.04}} & \vtop{\hbox{\strut 9.14}\hbox{\strut 6.70}\hbox{\strut 8.99}} & \vtop{\hbox{\strut 1.51}\hbox{\strut -0.12}\hbox{\strut 1.63}} & \vtop{\hbox{\strut 4.74}\hbox{\strut -0.56}\hbox{\strut 3.62}} & \vtop{\hbox{\strut 2.49}\hbox{\strut 1.39}\hbox{\strut 1.87}}\\
     I$_{0GCP}$  & \vtop{\hbox{\strut 21}\hbox{\strut 12}\hbox{\strut 9}} & \vtop{\hbox{\strut 4.37}\hbox{\strut 2.75}\hbox{\strut 7.93}} & \vtop{\hbox{\strut 14.66}\hbox{\strut 13.07}\hbox{\strut 18.62}} & \vtop{\hbox{\strut 1.61}\hbox{\strut -0.20}\hbox{\strut 2.41}} & \vtop{\hbox{\strut 5.42}\hbox{\strut -0.97}\hbox{\strut 5.65}} & \vtop{\hbox{\strut 4.06}\hbox{\strut 2.73}\hbox{\strut 5.17}}\\
     I$_{1GCP}$  & \vtop{\hbox{\strut 20}\hbox{\strut 12}\hbox{\strut 8}} & \vtop{\hbox{\strut 4.38}\hbox{\strut 2.02}\hbox{\strut 6.46}} & \vtop{\hbox{\strut 14.71}\hbox{\strut 9.63}\hbox{\strut 15.06}} & \vtop{\hbox{\strut 1.68}\hbox{\strut -0.01}\hbox{\strut 1.69}} & \vtop{\hbox{\strut 5.65}\hbox{\strut -0.06}\hbox{\strut 3.95}} & \vtop{\hbox{\strut 4.04}\hbox{\strut 2.02}\hbox{\strut 4.88}}\\
     I$_{2GCP}$  & \vtop{\hbox{\strut 23}\hbox{\strut 14}\hbox{\strut 8}} & \vtop{\hbox{\strut 4.35}\hbox{\strut 1.77}\hbox{\strut 11.08}} & \vtop{\hbox{\strut 15.43}\hbox{\strut 8.98}\hbox{\strut 25.44}} & \vtop{\hbox{\strut 2.13}\hbox{\strut 0.62}\hbox{\strut 2.08}} & \vtop{\hbox{\strut 7.55}\hbox{\strut 3.14}\hbox{\strut 4.77}} & \vtop{\hbox{\strut 3.79}\hbox{\strut 1.48}\hbox{\strut 8.96}}\\
    I$_{3GCP}$  & \vtop{\hbox{\strut 23}\hbox{\strut 14}\hbox{\strut 9}} & \vtop{\hbox{\strut 3.93}\hbox{\strut 1.22}\hbox{\strut 6.10}} & \vtop{\hbox{\strut 13.68}\hbox{\strut 6.23}\hbox{\strut 14.20}} & \vtop{\hbox{\strut 1.12}\hbox{\strut 0.14}\hbox{\strut 0.98}} & \vtop{\hbox{\strut 3.89}\hbox{\strut 0.72}\hbox{\strut 2.27}} & \vtop{\hbox{\strut 3.77}\hbox{\strut 1.20}\hbox{\strut 5.57}}\\
    \hline
\end{tabular} 
}
\label{tab:DBH_estimation_stats}
\end{center}
\end{table*}

\section{Discussion and Conclusions}\label{sec:Discussion}

\subsection{Autonomous flying}


The performance of autonomous navigation was promising in both test forests. The obstacle avoidance success rate was 100 \% in the 'Evo-medium' forest and 87.5 \% in the 'Evo-difficult' forest. However, the smoothness of the trajectories suffered from the increase in forest density. While in 'Evo-medium' 71.4 \% of the flights had a smooth trajectory, in 'Evo-difficult', at least one emergency stop was needed in all of the flights. However, apart from the one failed flight in 'Evo-difficult', the system was able to do a successful trajectory replanning and finish the mission after the emergency stops. Further consideration should be given to finding the optimal sensor for detecting thin and dry branches earlier. However, the current prototype already allows reliable and safe autonomous obstacle avoidance in easy and medium difficulty level boreal forests.   

The accuracy of the position estimate of VINS-Fusion was satisfactory for the short flights performed in this study, but during longer flights, the error would start to accumulate, resulting in high position errors. However, the standard deviation between the errors of the test flights was low and VINS-Fusion systematically estimated that the flight paths were shorter than in reality (as can be seen in Figure \ref{fig:vins_traj}). This indicates that a notable part of the position error could be eliminated by improving the system calibration. The results indicate that the estimates of VINS-Fusion were more accurate with online $t_d$ estimation than when the $t_d$ value was fixed to the one obtained from the calibration. However, further development of the prototype should include implementation of hardware-level synchronization between the camera and the IMU of the autopilot. 

Position errors can also be reduced by optimizing the estimate with loop closures, and the results from the walking tests show that VINS-Fusion can detect loop closures even in a homogeneous forest environment. Increasing the number of features tracked by VINS-Fusion increased the number of detected loops. However, increasing the number of features also increases the computational burden of the system, potentially disrupting real-time estimation. Finding the optimal value for the tracked features, as well as measuring the actual improvement of the loop closures to the position estimate, requires further studies.

\subsection{Accuracy of forest measurements}

The proposed methodology and the accuracy of the DBH estimates obtained were evaluated by comparing them with other relevant studies (Section \ref{sec:relatedEstimation}, Table
\ref{tab:reference_DBHestimation_stats}) considering the precision, autonomy of the system, weight, and complexity of the test environment. 

The highest estimation accuracies in the literature have been obtained in studies using either considerably heavier and larger ULS systems \citet{hyyppa2020under, hyyppa2020comparison, muhojoki24BENCHMARKING} operated in sparse forest environments or in environments with few reference trees \citet{shimabuku2023diameter}. The estimation results obtained in this study were close to those obtained in most other studies utilizing camera-based under-canopy drone systems \citep{kuvzelka2018mapping, he2025estimating360, krisanski2020photodrone}. Furthermore, other studies based on SfM \citep{shimabuku2023diameter, kuvzelka2018mapping, he2025estimating360, krisanski2020photodrone} had more complex flight paths that allowed the full reconstruction most of the tree trunks, whereas in this study, the trunks were only partially reconstructed from image data collected from a single flight line.

Considering the size of the systems, miniaturized lightweight systems were used in the studies by \citet{shimabuku2023diameter}, \citet{krisanski2020photodrone}, \citet{kuvzelka2018mapping}, \citet{prabhu2024treeFalcon}, \citet{cheng2024Treescope}, \citet{he2025estimating360}, and \citet{liang2024forest}, though the LiDAR-based custom drone systems \citep{prabhu2024treeFalcon, cheng2024Treescope, he2025estimating360, liang2024forest} were still considerably heavier and larger than the camera-based drone systems.
From these, LiDAR-based flights in \citet{cheng2024Treescope} and \citet{liang2024forest} were the only ones where the systems were reported to fly without commands from the pilot during the flight. Considering the complexity of the environments, the density of the forest in \citet{liang2024forest} (1000 trees/ha) was between the forests in this study (650 and 2000 trees/ha). In \citet{cheng2024Treescope}, the density of the forest was not reported.

In summary, the approach proposed in this study, camera-based autonomous under-canopy flight and photogrammetric point cloud processing from low-cost onboard stereo camera data, yielded a level of estimation close to the accuracy of most of the studies with SfM drone systems \citep{kuvzelka2018mapping, krisanski2020photodrone, he2025estimating360}. In particular, existing studies based on SfM of drones were using commercial drone systems piloted manually, except \citet{he2025estimating360} where the camera was added as an additional payload to a custom LiDAR-based drone platform, and the flight mode was not reported. Compared to studies with LiDAR-based miniaturized autonomous drone data collection, the accuracy of the estimation of DBH obtained in this study was higher than in \citet{liang2024forest} and close to that in \citet{cheng2024Treescope} even without any GCPs. However, the system utilized in \citet{cheng2024Treescope} has yielded slightly better accuracy with manually piloted flights \citep{cheng2024Treescope, prabhu2024treeFalcon}, suggesting that their LiDAR-based system has the potential for higher accuracy of estimation by optimizing autonomous flight trajectories.

Based on the results, the potential estimation accuracy with a low-cost stereo camera is considerably lower than that with heavy and expensive survey LiDARs but close to the quality of smaller LiDARs suitable for miniaturized drones capable of flying inside dense forests. However, the weight and price of the sensor are considerably lower, allowing the building of smaller platforms capable of flying in even denser forests, as well as reducing the cost of the system. The results show that GCPs can be used to improve the estimation quality (completeness and RMSE). However, the increase in estimation error was relatively low when GCPs were not used in postprocessing (RMSE from 12.98 \% to 14.66 \%) suggesting that the approach has the potential to work even without additional GCPs.  

Based on previous findings, it can be concluded that camera-based autonomous miniaturized drones have the potential to become a practical and cost-effective tool for forest parameter estimation in the future. However, further studies are required to increase the completeness of the reconstruction and the reconstruction coverage of the trees to decrease the RMSE of the estimation, especially for trees with large diameters.

\subsection{Potential improvements and future work}

The empirical evaluation of the system revealed areas for improvement, which can inform further development toward a practical tool for operational forest measurements.

The reliability of navigation and the smoothness of the trajectory could be further improved by investigating alternative camera systems and image processing methods (e.g., image super-resolution) with a better capability of detecting thin branches. Blind sides in the obstacle avoidance map could be tackled either by sensors with 360\textdegree{} FOV (e.g., 360\textdegree{} camera or multi-camera system) or by storing historical data in the occupancy grid map in addition to the current visible area. 

The virtual floor and ceiling of Ego-planner-V2 \citep{zhou2022swarm} are also defined as static with respect to the takeoff location of the drone, limiting low-altitude flying to relatively flat areas. To capture images at the DBH estimation altitude in uneven forest plots, the flight altitude limits should be modified to be dynamic with respect to the local terrain level.

The drift compensation remains also an important objective for future development. Initial loop closure tests yielded promising results, and future developments could assess the approach using block flights with a lawnmower pattern. In addition to the loop closures, the error in the position estimate could potentially be further reduced by fusing GNSS data to the VIO estimate. Potential methods for GNSS-Fusion include utilizing the global fusion features in VINS-Fusion \citep{qin2019global} or replacing VINS-Fusion with another method fusing GNSS-data and VIO, e.g., GVINS \citep{Cao2022GVINS}. Although the dense forest canopy blocks the GNSS signals, during long-term operations, the drone could receive corrections to the position estimate in the open areas while relying on the VIO estimate in denser parts of the forest.

It is also notable that the rigorous photogrammetric post-processing effectively compensated for the systematic errors. While the real-time trajectory error was 50 cm, the post-processed 3D error was approximately 11 cm without any GCPs.
Therefore, for applications relying on post-processing, drift is unlikely to present significant challenges.

In this study, the flight paths were straight flights through the sample plots, and the DBH values were estimated only for the tree trunks recorded during that path. For more comprehensive forest characterization, the trajectory should cover the whole test plot and observe the trees from multiple sides. In the study by \citet{he2025estimating360}, the DBH estimation error increased significantly when only part of the shape of the tree trunk was observed. This suggests that reconstructing the tree trunk from multiple sides instead of only one side could possibly also reduce the DBH estimate errors obtained with the system utilized in this study. In particular, this is expected to improve the quality of the estimation for trees of large diameter that yielded a higher RMSE in this study than trees with a DBH of less than 30 cm. The global path covering the entire test plot could be implemented with a simple waypoint generation algorithm, such as \citet{liang2024forest}, or with more advanced exploration methods such as those proposed by \citet{Zhou2021FUEL} or \citet{Cieslewski2017exploration}. Another way to achieve area coverage would be utilizing a swarm of drones, as in the original article by \citet{zhou2022swarm}. Furthermore, recent exploration algorithms \citep{Zhou2023RACER}\citep{Bartolomei2023swarmExploration} are able to utilize drone swarms to further improve and speed up the area coverage. In potential follow-up studies, the DBH estimation potential of the system could also be evaluated in alternative forest environments, for example, in sparser forest environments.

Although this study focused on demonstrating the potential for forest characteristics estimation through DBH estimation and tree detection, the scope of data collection extends well beyond these applications. Including an additional lightweight camera pointing upwards would enable a more comprehensive reconstruction of the tree trunks to perform also stem curve estimation based on the point clouds. With the developed camera-based prototype, data collection extends beyond point clouds to encompass applications such as image-based mapping and object detection. This opens up a variety of potential uses, including classifying and counting tree species and other vegetation, as well as monitoring bark damage caused by insect pests. 

The prototype also opens up possibilities for the study of real-time forest characteristics analysis. In this study, the point clouds were obtained through post-processing. However, Intel RealSense D435 is also capable of generating 3D point clouds in real time. Furthermore, image-based object detection can be performed in real time, for example, with YOLO \citep{Redmon2016yolo}. While only computationally light analysis can be conducted on the onboard computer, real-time analysis with a higher computational burden could be achieved through a remote computer or cloud service via an Internet connection.

\section*{Funding}

This research was funded by the Research Council of Finland within projects 'Learning techniques for autonomous drone based hyperspectral analysis of forest vegetation' (decision no. 357380), Fireman (decision no. 346710), and the Research Council of Finland Flagship Forest–Human–Machine Interplay—Building Resilience, Redefining Value Networks and Enabling Meaningful Experiences (UNITE) (decision no. 357908).

\section*{Declaration of Competing Interest}
The authors declare that they have no known competing financial interests or personal relationships that could have appeared to influence the work reported in this article.

\section*{Data Availability Statement}
Part of the data recorded during the field tests is available from the corresponding author upon reasonable request.

\section*{Declaration of AI-assisted technologies in the writing process}

During the revision of this work, the author(s) used free versions of Grammarly and Writefull (included in the Overleaf LaTeX editor) to improve the writing style in a non-native language manuscript. These tools were only used to detect grammar mistakes and improve the clarity of the text. All suggestions of these writing assistants were manually accepted by the author(s) and no text was autogenerated with generative AI tools. After using these tools, the author(s) reviewed the content and take(s) full responsibility for the content of the publication. 

\bibliographystyle{tfcad}
\bibliography{bibliography}

\end{document}